\documentclass{article}

\usepackage{arxiv}

\usepackage[utf8]{inputenc} 
\usepackage[T1]{fontenc}    
\usepackage{hyperref}       
\usepackage{url}            
\usepackage{booktabs}       
\usepackage{amsfonts}       
\usepackage{nicefrac}       
\usepackage{microtype}      
\usepackage{lipsum}
\usepackage{graphicx}
\usepackage{hyperref}
\usepackage{epstopdf, epsfig}
\usepackage{bm}
\usepackage{amsthm}
\usepackage{amsmath}
\usepackage{amssymb}
\usepackage[utf8]{inputenc}
\usepackage[T1]{fontenc}
\usepackage{mathptmx}
\usepackage{multirow}
\usepackage[numbers]{natbib}
\usepackage{float}
\usepackage{eucal}
\usepackage{color}
\usepackage{subfigure}
\usepackage[ruled,linesnumbered]{algorithm2e}
\usepackage{natbib}
\usepackage{caption}
\SetKwComment{Comment}{$\triangleright$\ }{}

\newcommand{\vect}[1]{\bm{#1}}

\newcommand{\matr}[1]{\mathbf{#1}}

\title{Deep Capsule Encoder-Decoder Network for Surrogate Modeling and Uncertainty Quantification}

\author{
  Akshay Thakur \\
  Department of Applied Mechanics\\
  Indian Institute of Technology Delhi\\
  Hauz Khas - 110016, New Delhi, India \\
  \texttt{akshaythakur1482@gmail.com} \\
   \And
 Souvik Chakraborty \\
  Department of Applied Mechanics\\
  School of Artificial Intelligence\\
  Indian Institute of Technology Delhi\\
  Hauz Khas - 110016, New Delhi, India \\
  \texttt{souvik@am.iitd.ac.in} \\
}

\begin{document}
\maketitle

\begin{abstract}
We propose a novel \textit{capsule} based deep encoder-decoder model for surrogate modeling and uncertainty quantification of systems in mechanics from sparse data. The proposed framework is developed by adapting Capsule Network (CapsNet) architecture into image-to-image regression encoder-decoder network. Specifically, the aim is to exploit the benefits of CapsNet over convolution neural network (CNN) $-$ retaining pose and position information related to an entity to name a few. The performance of proposed approach is illustrated by solving an elliptic stochastic partial differential equation (SPDE), which also governs systems in mechanics such as steady heat conduction, ground water flow or other diffusion processes, based uncertainty quantification problem with an input dimensionality of $1024$. However, the problem definition does not the restrict the random diffusion field to a particular covariance structure, and the more strenuous task of response prediction for an arbitrary diffusion field is solved. The obtained results from performance evaluation indicate that the proposed approach is accurate, efficient, and robust.  
\end{abstract}


\section{Introduction}\label{S:1}
The lack of complete knowledge about a system in mechanics or some randomness intrinsic to the system leads to emergence of uncertainty in numerical simulators. In order to ascertain the effect of such uncertainty on the output of a numerical simulators, it is essential that one looks towards the field of uncertainty quantification \cite{smith2013uncertainty}. More precisely, it is only relevant the problem under consideration be reformulated into an uncertainty propagation (UP) problem.\par
The most straightforward way to solve UP problem is via the usage of Monte Carlo (MC) method \cite{mooney1997monte}, which requires considerably large number of repeated evaluation of the solution of the problem at random input samples for getting convergent statistics. Now, of course, if we take into account the high computational cost of a single simulation run for complex multiscale and multiphysics systems, the situation becomes even more daunting and expensive for repeated evaluations. Furthermore, even the more advanced techniques such as Latin Hypercube Sampling \cite{loh1996latin} and Quasi MC-method \cite{caflisch1998monte} do not come to rescue and are often not apt for UP problems. Therefore, in such scenarios, the rational choice is to construct computationally efficient surrogate models (SM) which could then be queried instead of the original simulator using sampling methods such as the MC-method for completing UQ tasks. Further, some of the notable approaches for surrogate constructions in literature include polynomial chaos expansion \cite{blatman2011adaptive, xiu2002wiener, elsheikh2014efficient}, Gaussian processes \cite{williams2006gaussian, bilionis2012multi,chakraborty2019graph,pandita2021surrogate}, variance decomposition analysis \cite{chakraborty2017efficient,chakraborty2017polynomial} and its variants \cite{chakraborty2017moment}, support vector machines \cite{roy2019support}, and deep neural networks \cite{sun2020surrogate,pawar2022hyperparameter,papadopoulos2018neural,zhang2021multi,haghighat2021physics,thakur2021deep}.\par
The disadvantages of most of these surrogate modeling approaches barring deep neural networks is their intractability when the dimensionality of the problem at hand becomes high. The most commonly used approaches in the field of uncertainty quantification to tackle this issue is utilisation of dimensionality reduction techniques such as Karhunen-
Loève expansion (KLE) \cite{ghanem2003stochastic}, kernel principal component analysis \cite{ma2011kernel} or proper orthogonal decomposition \cite{jacquelin2019random}, and active subspaces \cite{constantine2014active,constantine2015exploiting,navaneeth2022surrogate}. However, the issue with dimensionality reduction techniques is inefficiency, as suitable latent representations have to found before building the regression model. Nevertheless, concomitantly, it is also true that deep neural networks offer a dependable alternative because of their nonlinear function approximation capabilities and their ability to scale to high dimensions \cite{chakraborty2021transfer,zhu2018bayesian,yang2019conditional}.\par
Previous studies in the literature such as Tripathy and Bilionis \cite{tripathy2018deep} developed a multi-layer perceptron based approach for constructing excellent surrogate models for SPDE-governed systems in mechanics. However, the approach requires a considerable grid-search and Bayesian optimisation for finding an optimal network structure, and also, require a large number of epochs for successful training of the model. In addition, \cite{zhu2018bayesian,khoo_lu_ying_2021, mendu2021gated, zhu2019physics} have used convolutional neural network based approaches for uncertainty quantification and surrogate modeling and have produced remarkable results. However, in these studies, either a large number of training samples are required for achieving reasonable accuracy or there is need for considerable cross-validation for arriving at a satisfactory network structure. Furthermore, novel architectures such as CapsNet \cite{sabour2017dynamic} have been introduced which address several shortcomings of CNNs such as their inability to learn the relative spatial relationships between two entities or the precise location and pose of a particular entity in an image. Furthermore, it has also been found that Capsule Networks perform better than CNNs when the size of training dataset is small while providing a comparable performance for larger training datasets \cite{mobiny2019automated}. Therefore, to realize the benefits of these advantages, in this study, we develop a novel end-to-end image-to-image regression framework based on CapsNet for uncertainty quantification and surrogate modeling of systems in mechanics. Additionally, we also intend this framework to be able to successfully tackle high dimensional problems, be efficient in training, and be robust to noise and limited data availability.\par
The remainder of the paper is organised as follows. Section \ref{S:2} provides details on the problem statement. In Section \ref{S:3}, we introduce Capsule Networks, describe how we adapt them to accomplish our end-to-end image-to-image regression task, and lay forth the procedure for training the network. Section \ref{S:40} contains details about the numerical problem to be solved and presents the performance results of the proposed framework on the same problem. The numerical problem is SPDE system which governs systems in mechanics such as steady state heat conduction, ground water flow and other diffusion processes \cite{xiu2002modeling,lykkegaard2021accelerating,li2016inverse}. However, we follow the problem definition similar to Tripathy and Bilionis \cite{tripathy2018deep}, where no assumptions are placed on uncertain diffusion's regularity and lengthscale. Also, when defined in this way, it is not possible to perform dimensionality reduction such as KLE beforehand, which makes the problem substantially more difficult. Finally, the concluding remarks are provided in Section \ref{S:6}.   
\section{Problem Statement}\label{S:2}
Consider a stochastic partial differential equation (SPDE) on an $m$-dimensional domain, $D \subset \mathbb{R}^{m}$, bounded by $\partial D$ of the following form
\begin{equation}
\mathcal{M}(\vect{s},\vect{x}, \omega ; y)=0, \forall \vect{x} \in D,
\label{equation1}
\end{equation}
with boundary conditions:
\begin{equation}
\mathcal{B}(\vect{s},\vect{x}, \omega ; y)=0, \forall \vect{x} \in \partial D,
\label{equation2}
\end{equation}
where $s \in \vect{S} \subset \mathbb{R}^{m}$ are the physical coordinates or the spatial locations in $m$-dimensional domain $D$ where the solution is evaluated,  $\omega$ is an  element and an event in the event space $\Omega$ of the probability space $\{\Omega,\Sigma, \mathbb{P}\}$, $\vect{x(s, \omega)} \in \mathbb{R}^{\zeta_{x}}$ is the random field, $y(s,x(s)) \in \mathbb{R}^{\zeta_{y}}$  is the SPDE's response, $\mathcal{M}$ denotes a partial differential operator, and the boundary operator enforcing the boundary conditions is denoted with $\mathcal{B}$. In the present study, the residence of our interest is in SPDE-governed physical systems. More specifically, we focus on the computer simulation of the systems modeled by SPDE which are solved on the set of physical coordinates or locations on a spatial grid $\vect{s} = (s_{1}, s_{2},..., s_{n})$ with input being the random field $\boldsymbol{\mathrm{x}} \in \mathcal{X} \subset \mathbb{R}^{n\zeta_{x}}$ and the output response or the solution being  $\boldsymbol{\mathrm{y}} \in \mathcal{Y} \subset \mathbb{R}^{n\zeta_{y}}$, where $\zeta_{x}$ and $\zeta_{y}$ are the input and output channels respectively. The simulator could also be represented as the following function: $f: \mathcal{X} \mapsto \mathcal{Y}$. Considering the spatial grid to be two-dimensional with $n_{x}$ and $n_{y}$ being the number of grid points in the respective grid axes directions, the above-mentioned mapping could be re-formulated into a image regression task, which could be  represented as follows
\begin{equation}
f: \mathbb{R}^{n_{x} \times n_{y} \times \zeta_{x}} \mapsto \mathbb{R}^{n_{x} \times n_{y} \times \zeta_{y}}
\label{equation3}
\end{equation}
Therefore, naturally, the task of surrogate modeling could be performed by training on an image dataset $\mathcal{D}_{XY}=\left\{\boldsymbol{\mathrm{x}}_{i}, \boldsymbol{\mathrm{y}}_{i}\right\}_{i=1}^{N}$, where $\boldsymbol{\mathrm{x}}_{i} \in \mathbb{R}^{n_{x} \times n_{y} \times \zeta_{x}}$ is one instance of the input random field, $\boldsymbol{\mathrm{y}}_{i} \in \mathbb{R}^{n_{x} \times n_{y} \times \zeta_{y}}$ is a single instance of the output field, and $N$ is the number of times a simulation is ran or the number of training images.\par
It is well known that conducting statistical studies such as uncertainty quantification often times becomes quite computationally expensive as it requires repeated runs of the simulator, and this is what justifies the need for surrogate models in order to make such statistical tasks inexpensive. It is also true that a significant amount of work has been performed in the research sphere of surrogate modeling including image-to-image regression, deep learning-based surrogate models. However, the previous deep-learning based models either require a considerable amount of cross-validation or a large number of training runs to reach at an optimal result. Therefore, in this paper, our aim is develop an image-to-image regression and deep learning based surrogate model based on a newly deep learning architecture which addresses both of the aforesaid shortcoming and offers additional benefits.
\section{Methodology}\label{S:3}
\subsection{Capsule Networks}\label{S:31}
\subsubsection{Inputs and outputs of a capsule and dynamic routing}\label{S:311}
As defined by Sabour et al. \cite{sabour2017dynamic}, a capsule is a collection of neurons with their activity vectors describing the various properties like pose, texture, velocity etc of a particular entity or a part of it in an image, where the properties are also the instantiation parameters for that entity. The output of a capsule is a vector whose overall length represents the probability of the instantiated entity's existence in the image while its orientation represents the instantiation parameters. Also, it is important to note that the a nonlinear function also called a squashing function is applied on the output vector of the capsules which although does not alter the orientation of the vector, but guarantees that the overall length of the vector remains between 0 and 1. Mathematically, the squashing function is represented as follows
\begin{equation}
\mathbf{l}_{j}=\frac{\left\|\mathbf{p}_{j}\right\|^{2}}{1+\left\|\mathbf{p}_{j}\right\|^{2}} \frac{\mathbf{p}_{j}}{\left\|\mathbf{p}_{j}\right\|}
\label{equation4}
\end{equation}
where the vector output of $j^{th}$ capsule is represented by $\mathbf{l}_{j}$ while the total input to the same capsule is represented by $\mathbf{p}_{j}$.
Furthermore, due to vectorial nature of the capsule output Sabour et al. \cite{sabour2017dynamic} introduced a dynamic routing mechanism which helped in making sure that an appropriate parent in the subsequent layer receives the output from a given capsule. The mechanism is implemented by firstly obtaining prediction vectors $\hat{u}_{j \mid i}$ from first capsule layer onward by multiplying the weight matrix $\mathbf{W}_{i j}$ of the capsule in the layer under consideration with the output $\mathbf{u}_{i}$ of the same capsule $(\hat{\mathbf{u}}_{j \mid i}=\mathbf{W}_{i j} \mathbf{u}_{i})$. Then the input $\mathrm{p}_{j}$, which is routed to all the capsules in the subsequent layer, is obtained by performing the weighted sum over all prediction vectors  $\hat{u}_{j \mid i}$ from the previous layer, and mathematically, is represented by the following equation
\begin{equation}
\mathrm{p}_{j}=\sum_{i} c_{i j} \hat{\mathbf{u}}_{j \mid i}, 
\label{equation5}
\end{equation}
where $c_{i j}$ are the coupling coefficients which are learned iteratively through a dynamic routing procedure. The sum of the coupling coefficients between capsule $i$ and every capsules in the subsequent layer evaluates to 1. These coefficients are ascertained with the help of following expression
\begin{equation}
c_{i j}=\frac{\exp \left(b_{i j}\right)}{\sum_{k} \exp \left(b_{i k}\right)}, 
\label{equation6}
\end{equation}
and the $\log$ prior probabilities of a capsule $i$ coupling to a capsule $j$ are set as the initial values of logits $b_{i j}$.  The coupling coefficients are then updated in an iterative fashion based on the scalar product (also referred as agreement in current scenario) between  the output $\mathbf{l}_{j}$ of every capsule $j$ in a given layer and the prediction $\hat{\mathbf{u}}_{j \mid i}$ made by capsule $i$ in the preceding layer, which could be represented as follows 
\begin{equation}
a_{i j}=\mathbf{l}_{j} . \hat{\mathbf{u}}_{j \mid i} . 
\label{equation7}
\end{equation}
 The resulting agreement value obtained is then added to  $b_{i j}$, and is ensured by computation to update the values for each coupling coefficients between capsule in corresponding layers. It so happens that with each subsequent routing iteration the coupling coefficients between a capsule $i$ in a given layer and an appropriate parent capsule $j$ in subsequent layer increases along with the increase in contribution of capsule $i$ to output of capsule $j$. 
 \subsubsection{A simple Capsule Networks Architecture}\label{S:312}
\begin{figure}[H]
\includegraphics[width=1\textwidth]{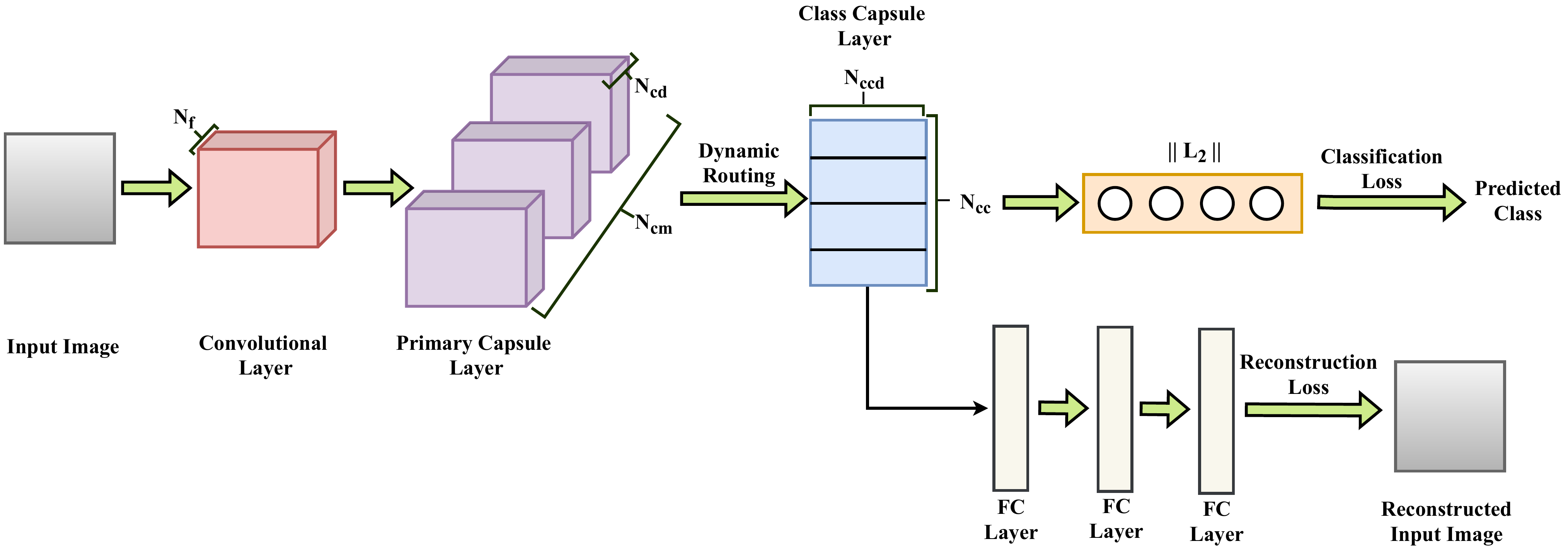}
\caption{A simple three layer CapsNet augmented with a FCN decoder for tasks of classification and reconstruction}
\label{fig:0}
\end{figure}

 In Figure \ref{fig:0}, a simple capsule network similar to the one proposed in \cite{sabour2017dynamic} is presented. The first layer of the network is a convolutional layer which extracts features from the input image, which are then used as an input to the subsequent Primary Capsule layer. The Primary Capsule layer comprises of  $N_{cm}$ number of convolutional capsule channels or maps with the number of dimensions of each convolutional capsule's output vector being $N_{cd}$.  Also, it is simultaneously true that the number of convolutional maps in the network's first layer $N_{f} = N_{cm}\times N_{cd}$. It is also key to note that the activation of a primary capsule is commensurate to the inversion of graphic rendering process. The final layer of the part of network responsible for classification is the class capsule layer (also, referred to as digit capsule layer in \cite{sabour2017dynamic}). This layer contains $N_{cc}$ number of capsules with $N_{ccd}$ number of dimensions, where $N_{cc}$ is equal to the number of different classes in the dataset which is to be classified. Furthermore, the process of dynamic routing only takes place between the primary capsule and the class capsule layer. The outputs of class capsule layer are then subject to the $L_{2}$ norm and classification predictions are then obtained by training the network using a suitable classification loss.  
As shown in Figure \ref{fig:0}, Sabour et al. \cite{sabour2017dynamic} augmented the classification network with a Fully Connected Network (FCN) for reconstruction of the input image by using mean squared loss. The FCN also served as a regularisation method and promoted the learning of entity instantiation parameters' proper encoding in the class capsules.   

In drawing a comparison between CNNs and CapsNet, we note that, like CNNs, CapsNet also has the ability to translate knowledge from one position on an image to other. However, unlike CNNs scalar-output feature detectors, CapsNet have vector-output capsules. Also, the routing process, routing-by-agreement, used in CapsNet is much more advanced than the other basic forms of routing such as max pooling used in CNNs, and it helps in retaining the information about the position and orientation of an entity in an image as compared to max pooling's retaining of only the information about entity's existence. Furthermore, an additional advantage of CapsNet architecture is the ability of higher-level capsules to represent entities which are considerably more complex with greater number of degrees of freedom depending upon the capsule's dimensionality. These are reasons which justify the need of exploration of CapsNet architecture for the tasks of image regression.
\subsection{Proposed Framework}\label{S:32}
The idea behind encoder-decoder networks is to first detect high-level features in the image by reducing the spatial dimensions using the encoder and then, to get back the spatial dimensions from these detected features using the decoder. However, in case of image-to-image regression task, it needs to be noted that the recovered or the output image could be vastly different from the input image. We also follow an encoder-decoder network approach and adapt the CapsNet architecture into an extended image-to-image regression encoder-decoder network. Essentially, our proposed network consists of a CapsNet-based encoder, a dense layer decoder, and a transposed convolution layer or Transposed convolutional decoder.

\subsubsection{CapsNet-based encoder}\label{S:321}
\begin{figure}[H]
\centering
\includegraphics[width=0.7\textwidth]{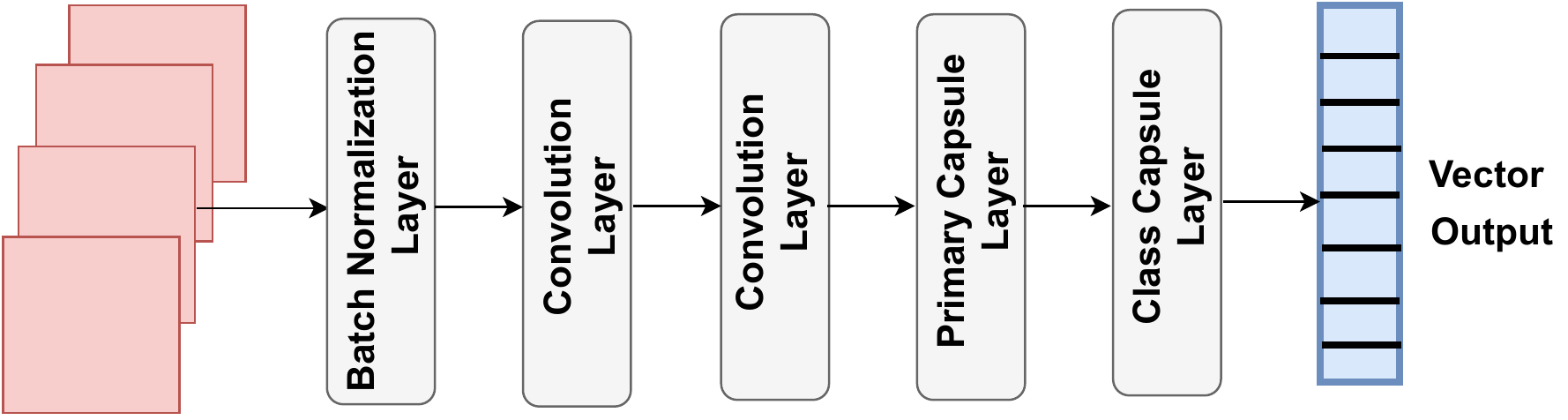}
\caption{A CapsNet encoder with two convolution layers, a primary capsule layer, and a class capsule layer}
\label{fig:1}
\end{figure}

As represented in Figure \ref{fig:1}, we have performed slight modifications to CapsNet encoding architecture presented in Section \ref{S:312} and have prepended the said architecture with additional convolutional layer and a batch normalisation layer to arrive at the encoding network we would be utilising. The network encodes the input images into a total of $N_{cc}$ vectors with $N_{ccd}$ number of dimensions.   

\subsubsection{Dense layer decoder}\label{S:322}
\begin{figure}[H]
\centering
\includegraphics[width=0.7\textwidth]{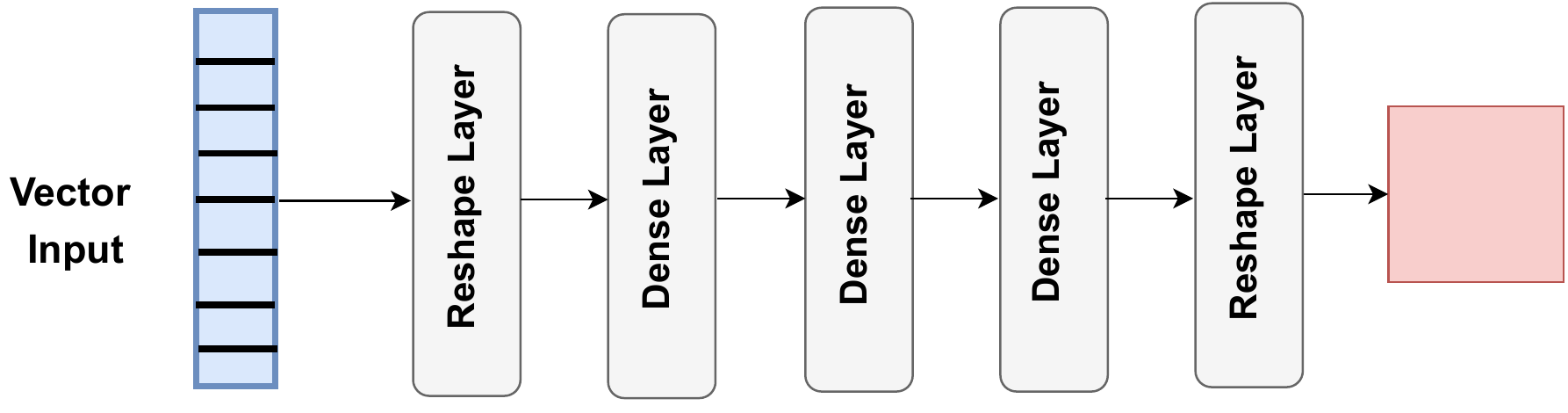}
\caption{A dense layer decoder consisting of three dense layers and two reshaping layers.}
\label{fig:2}
\end{figure}

To decode the information which is encoded in the form of vectors by the CapsNet encoder and to reshape it back into an image, we use a decoder, similar to one used for reconstruction in Section \ref{S:312}, comprising of multiple fully connected dense layers and reshaping layers. The dense layer decoder network is presented in Figure \ref{fig:2}.

\subsubsection{Transposed convolutional decoder}\label{S:323}
\begin{figure}[H]
\centering
\includegraphics[width=0.7\textwidth]{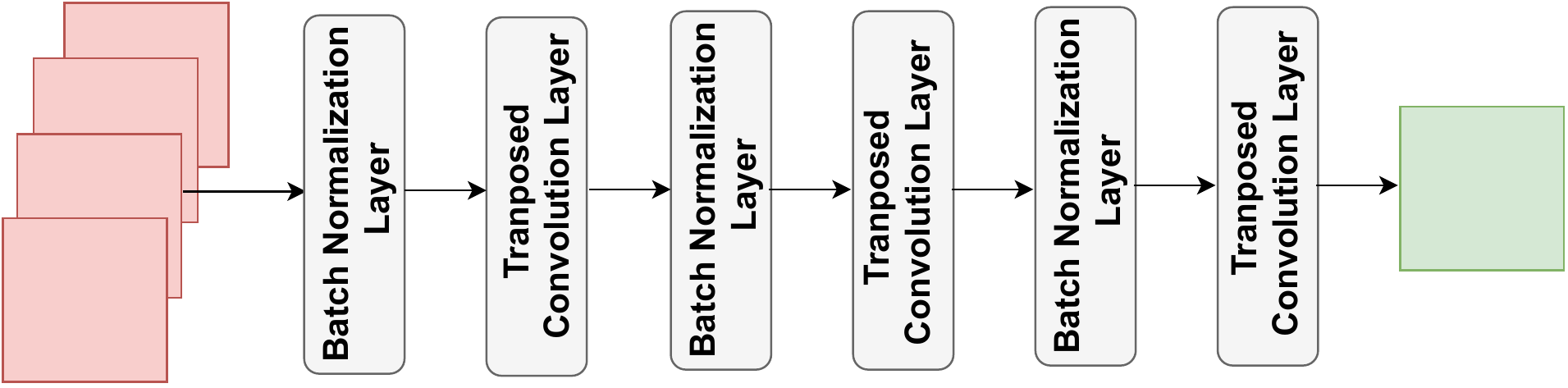}
\caption{A transposed convolutional decoder with three subsequent sets of batch normalization and transposed convolution layers}
\label{fig:3}
\end{figure}

Finally, we propose the use of an additional decoder which comprises of three consecutive sets of batch normalization and transposed convolution layers and is presented in Figure \ref{fig:3}. 
The transposed convolution layers are configured such that the image size the same after passing through the layer. However, we use a larger number of feature maps in the first transposed convolution layer of this decoder and then the number of channels is reduced to one towards the final layer.

\subsubsection{Complete Framework}\label{S:324}
\begin{figure}[H]
\includegraphics[width=1\textwidth]{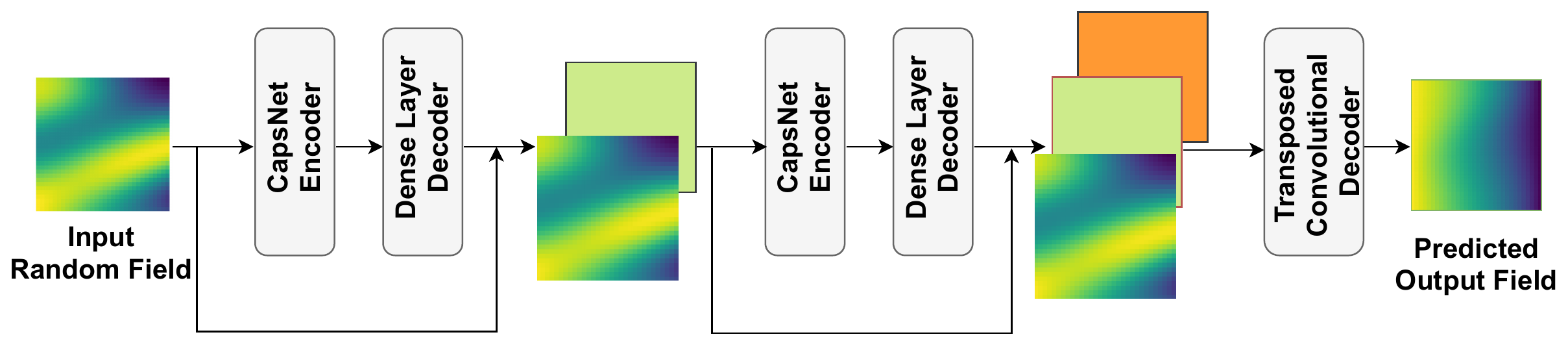}
\caption{The proposed CapsNet based image-to-image regression network with two subsequent sets of CapsNet encoder and dense layer decoder followed by the transposed convolutional decoder. Also between subsequent sets of Capsule encoder and dense layer decoder skip connections are established.}
\label{fig:4}
\end{figure}

By utilising the encoder and decoders described in the sections above, we develop a CapsNet based framework for image-to-image regression which is illustrated in Figure \ref{fig:4}. The input image or field is fed into the CapsNet encoder and the resulting encoded vectorial output is then provided to the dense layer decoder. Furthermore, a skip connection is established by concatenating the output of the dense layer decoder, which is an image with the same dimensions as the input image, with the input image. The underlying usage of this step is to ensure that the continuous connection between the input and output image is maintained so that any direct effects of input image on the output image could be modeled. After this, the network is once again augmented with the similar block of CapsNet encoder, dense layer decoder and a skip connection. Finally, the output from this block is passed through the transposed convolutional decoder and the resulting image or field is the final output from network, which can also be represented as $\boldsymbol{\mathrm{y}}=f(\boldsymbol{\mathrm{x}; \theta)}$, where $\boldsymbol{\theta}$ are the model parameters.  
\subsection{Training}\label{S:325}
The propounded CapsNet based surrogate model has been implemented in TensorFlow \cite{tensorflow2015-whitepaper}. We use mean squared error (MSE) as the loss function for our network. The number of dynamic routing iterations between the primary and class capsule layer in the CapsNet encoded is set to 3. The number of convolutional capsule feature maps $N_{cm}$ and output dimension of capsules $N_{cd}$ in the primary capsule layer of the encoder is set to 32 and 8, respectively. The number of capsules $N_{cc}$ in the class capsule layer of the encoder and their output vector's dimension $N_{ccd}$ is made equal to 30 and 16, respectively. Automatic gradient calculation is performed using the available TensorFlow functionalities and the Adam optimizer is employed. The initial learning rate is set to 0.05 and a custom learning rate schedule is implemented for proper training. In addition, the algorithm for training the network is presented in Algorithm \ref{alg:one} for reader's perusal. The complete source code for the framework would be made available on GitHub on publication of this work and the corresponding link would be provided.

\begin{algorithm}[H]
\caption{Training of CapsNet based image-to-image regression surrogate}\label{alg:one}
\textbf{Input:} Training dataset  ${\mathcal{D}} = \left\{\left(\boldsymbol{\mathrm{x}}_{i}, \boldsymbol{\mathrm{y}}_{i}\right)\right\}_{i=1}^{N}$ is provided\\
Dataset ${\mathcal{D}}$ is split into ${\mathcal{D}}_{train}$, ${\mathcal{D}}_{val}$, and ${\mathcal{D}}_{test}$.\\ 
Divide ${\mathcal{D}}_{train}$ into $N_{m}$ mini batches of size $b = N/N_{m}$ randomly\\
 \For{$e= 1$ to  $number\; of\; epochs$}    
        { 
        	 \For{$n_{m} = 1$ to $N_{m}$}    
                { 
                	Draw a mini batch ${\mathcal{A}}$ from $\boldsymbol{\mathcal{D}}$\\
                	 \For{$l= 1$ to $b$}    
                        { 
        	              $\boldsymbol{\mathrm{y}}_{l}\leftarrow f(\boldsymbol{\mathrm{x}}_{l} ; \boldsymbol{\theta})$
                        }
        	        Store all $(\boldsymbol{\mathrm{x}}_{l},\boldsymbol{\mathrm{y}}_{l})$ in dataset ${\mathcal{A}}^{\prime}$\\
        	        Calculate the gradient $\frac{\partial MSE}{\partial \theta}$ on mini batch datasets $\mathcal{A}$ and $\mathcal{A}^{\prime}$.\\
        	        Update $\boldsymbol{\theta}$ using gradient-based Adam optimizer
                }
        }
\textbf{Output:} Samples of response $\boldsymbol{\mathrm{y}}_{j},\,j=1,2,...,N$
\end{algorithm}
\section{Numerical Implementation and Results}\label{S:40}
\subsection{Problem definition}\label{S:4}
We  consider the following two-dimensional stochastic elliptic PDE, which models a  steady state diffusion process, on a two-dimensional Euclidean domain bounded by unit square 
\begin{equation}
-\nabla\cdot(a(\boldsymbol{\mathrm{x}}) \nabla u(\boldsymbol{\mathrm{x}}))=0, \quad \forall \boldsymbol{\mathrm{x}} \in[0,1]^{2} 
\label{equation8}
\end{equation}
where $a(\boldsymbol{\mathrm{x}})$ is the diffusion coefficient varying in the two-dimensional space. Also, the following boundary conditions are enforced on the equation:
$$
\begin{array}{r}
u=0, \forall x=1, \\
u=1, \forall x=0, \\
\frac{\partial u}{\partial n}=0, \forall y=0 \text { and } y=1,
\end{array}
$$
where $\boldsymbol{\mathrm{x}}=(\mathrm{x}_{1}, \mathrm{x}_{2})$ are the spatial grid points.\par
In this numerical example, the diffusion coefficient $a$ is modeled as a log normal random field in order to capture the uncertainty associated with it. Mathematically, the same is expressed in the following way
\begin{equation}
\log a(\mathbf{x}) \sim \operatorname{GP}\left(a(\mathbf{x}) \mid m(\mathbf{x}), k\left(\mathbf{x}, \mathbf{x}^{\prime}\right)\right), 
\label{equation9}
\end{equation}
where $m(\mathbf{x})$ is the mean function and $k\left(\mathbf{x}, \mathbf{x}^{\prime}\right)$ is the covariance function of a Gaussian random field. 
For the current example, the mean is set to zero, i.e., $m(\mathbf{x}) = 0$. Also, the covariance function is modeled using an exponential kernel such that
\begin{equation}
k\left(\mathbf{x}, \mathbf{x}^{\prime}\right)=\exp \left(-\sum_{i=1}^{2} \frac{\left|x_{i}-x_{i}^{\prime}\right|}{\ell_{i}}\right), 
\label{equation10}
\end{equation}
where the correlation length in $i^{th}$ direction in space is denoted by $\ell_{i}$. Furthermore,  following the problem definition similar to Tripathy and Bilionis \cite{tripathy2018deep}, rather than building a surrogate model for only one length scale $\ell_{i}$, we also let go of constraints on $\ell_{i}$ and construct a surrogate model capable of predicting the solution $u$ at an random input or diffusion field $a(\mathbf{x})$  based on any arbitrary length scale kernel.  
\subsection{Data Generation}
We discretize a unit square domain into a finite volume (FV) grid $32 \times 32$ cells and utilise the finite volume method (FVM) Python library \textbf{FiPy} \cite{4814978} to solve the numerical example under consideration. The diffusion coefficient $a$ is modeled as a log-normal random field over the grid $\matr{s}$ to form the input $\hat{\mathbf{a}} \in \mathbb{R}^{32 \times 32}$ for the forward solver while $\hat{\mathbf{u}} \in \mathbb{R}^{32 \times 32}$ represents the solution output from the solver. Relegating to our previous definitions in Section \ref{S:2}, we let  $\matr{s}$ to be a set of cell center locations in 2-d $32 \times 32$ spatial grid, i.e., $\matr{s}=\left\{s_{1}, \ldots s_{n}\right\}$, $\hat{\mathbf{a}}\in \mathcal{X} \subset \mathbb{R}^{n\zeta_{x}}$ to be the input to the surrogate model, and $\hat{\mathbf{u}}\in \mathcal{X} \subset \mathbb{R}^{n\zeta_{y}}$ to be the surrogate model's output. Therefore, surrogate model's objective is to learn the mapping $\mathcal{X} \mapsto \mathcal{Y}$. Also, here, $\zeta_{x} = \zeta_{y} =1$ and $n = 32\times32$. Therefore, on reformulating our current problem description, according to our previous definitions in Section \ref{S:2}, into an image-to-image regression problem, we have
\begin{equation}
f: \mathbb{R}^{32 \times 32 \times 1} \mapsto \mathbb{R}^{32 \times 32 \times 1}. 
\label{equation11}
\end{equation}
We utilise the strategy for generating the realizations of diffusion coefficient $a$ proposed by Tripathy and Bilionis \cite{tripathy2018deep}. Also, the strategy is biased towards choosing smaller length scales diffusion fields, and thus, serves the purpose of realising the maximum variability in the input random fields. Also, in a similar fashion to the said study, we restrict the value of $\ell_{i}$ to be larger than the size of the FV cell and we set the lower bound $h_{l}$ to $\frac{1}{31}$. Following the mentioned restrictions and strategy, $n_{l}$ distinct pairs of lengthscale in $x$ and $y$ direction, $(\ell_{x}, \ell_{y}) \in \boldsymbol{\mathcal{L}}$ are generated. In addition, for every lengthscale pair, $r_{l}$ samples of diffusion coefficient $a$ are also generated. Also, Algorithm \ref{alg:two} provides the description of data generation procedure. Furthermore, for the purposes of current study, $n_{l} = 40$ and $r_{l}=50$.\par

\begin{algorithm}[H]
\caption{Generation of data}\label{alg:two}
\textbf{Require:} Lower bound on length scale $h_{l}$, the number of distinct lengthscale pairs $n_{l}$, the number of realizations for each lengthscale pair $r_{l}$\\
\textbf{Initialize:} Counter $b = 1$ and an empty array for lengthscale pairing storage $\boldsymbol{\mathcal{L}}$. \\ 
\While{$d\leq40$}{
Sample $(\alpha_{1}, \alpha_{2}, \alpha_{3}) \sim \mathcal{U}([0,1]^{3})$\\
\If{$exp(-\alpha_{1}-\alpha_{2} < \alpha_{3})$}{
    Set $\ell_{b}=\left(\ell_{x, b}, \ell_{y, b}\right)=\left(h_{l}+\alpha_{1}(1-h_{l}), h_{l}+\alpha_{2}(1-h_{l})\right)$\\
    Set $\boldsymbol{\mathcal{L}_{b}} \leftarrow \ell_{b}$\\
    Increment the counter $b\leftarrow b+1$
}
}
\For{$\left(\ell_{x}, \ell_{y}\right) \in \boldsymbol{\mathcal{L}}$}    
        { 
               Generate $r_{l}$ samples of diffusion coefficient $a$ having lengthscales $\ell_{x}$ and $\ell_{y}$\\
               Obtain the solutions for each $a$ using the FV Solver
        }
\end{algorithm}
\vspace{2em}

\begin{figure}[H]
    \centering
    \includegraphics[width=0.9\textwidth]{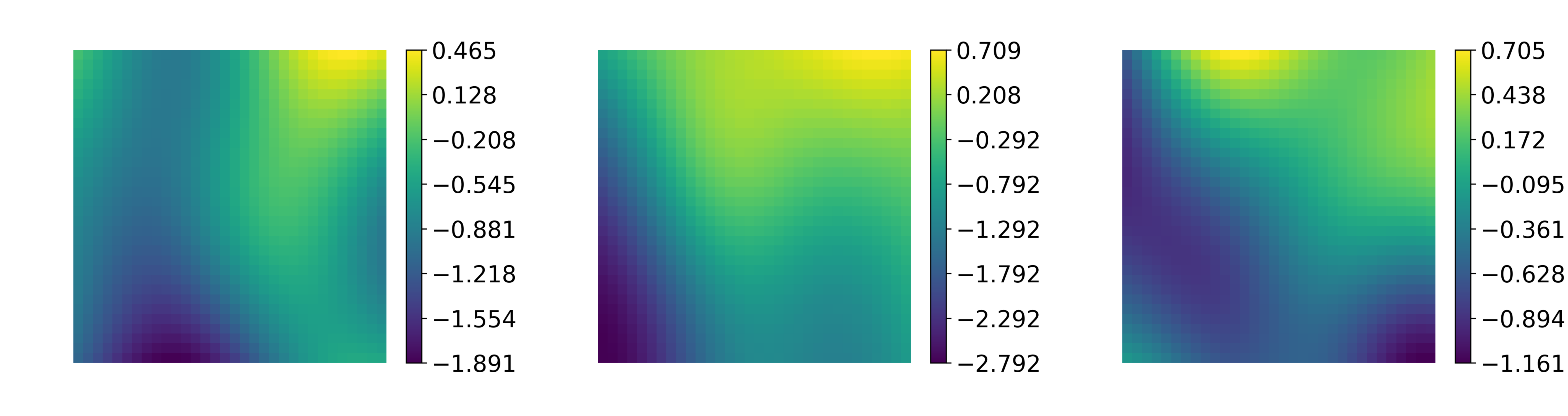}
    \caption{Three samples of logarithm of random diffusion field $\log a(\mathbf{x})$ for lengthscale pair $\ell_{x} = 0.3600$ and $\ell_{y} = 0.5668$.}
    \label{fig:5}
\end{figure}

\begin{figure}[H]
\centering
\includegraphics[width=0.75\textwidth]{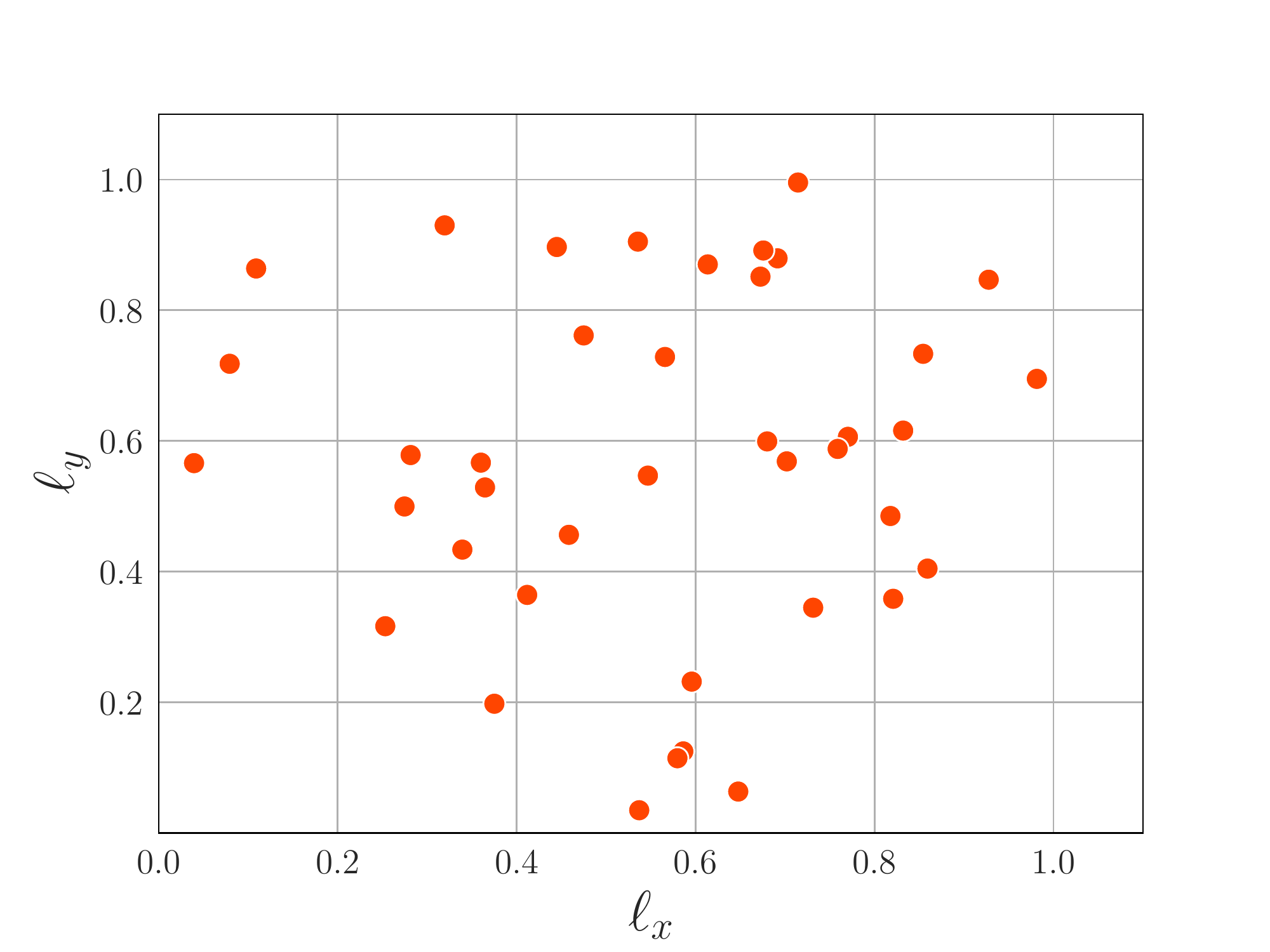}
\caption{Scatterplot of lengthscale pairs generated using Algorithm \ref{alg:two} for training the CapsNet-based image-to-image regression framework.}
\label{fig:6}
\end{figure}

Three samples of $\log a(\mathbf{x})$ for lengthscale $\ell_{x} = 0.3600$ and $\ell_{y} = 0.5668$ are shown in Figure \ref{fig:5}. Also, the $40$ lengthscale pairs generated for the current study are also presented using a scatter plot in Figure \ref{fig:6}. Finally, the dataset $\mathcal{D}$ containing $n_{l}\times r_{l}$, i.e, $40\times50 = 2000$, different pairs of $\hat{\mathbf{a}}$ and $\hat{\mathbf{u}}$ generated data using Algorithm \ref{alg:two} is split into training dataset $\mathcal{D}_{train}$, validation dataset $\mathcal{D}_{val}$, and test dataset $\mathcal{D}_{test}$. Also, the number of pairs of $\hat{\mathbf{a}}$ and $\hat{\mathbf{u}}$ in $\mathcal{D}_{train}$, $\mathcal{D}_{val}$, and $\mathcal{D}_{test}$ are $1000$, $500$, and $500$, respectively.  
\subsection{Results and Discussion}\label{S:5}
In this section, the performance of the proposed approach is demonstrated by constructing surrogate for numerical example defined in Section \ref{S:4}. In addition, to evaluate the accuracy of the propounded approach, we use two metrics for quantification of error between the predicted and actual solution, specifically, mean-squared error and  coefficient of determination ($R^{2}$-score). Firstly, the mean-squared error is evaluated as follows
\begin{equation}
L\left(\hat{\mathbf{u}}_{sm}, \hat{\mathbf{u}}_{FV}\right) = \frac{1}{1024}\left(\,\sum_{j=n_{1}}^{1024}\left(\hat{\mathbf{u}}_{sm, j}-\hat{\mathbf{u}}_{FV, j}\right)^{2}\right),
\label{equation12}
\end{equation}
where $j$ denotes the index of cell centers at which solution is obtained using FVM, $\hat{\mathbf{u}}_{\mathrm{sm},j}$ denotes the solution predicted by the proposed surrogate model  and $\hat{\mathbf{u}}_{\mathrm{FV},j}$ denotes the solution obtained using FVM at $j^{th}$ cell centers for a given diffusion field realization â. Secondly, the $R^{2}$-score is computed as follows
\begin{equation}
R^{2}=1-\frac{\sum_{j=1}^{1024}\left(\hat{\mathbf{u}}_{\mathrm{FV}, j}-\hat{\mathbf{u}}_{\mathrm{sm}, j}\right)^{2}}{\sum_{j=1}^{1024}\left(\hat{\mathbf{u}}_{\mathrm{FV}}-\overline{\mathbf{u}}_{\mathrm{FV}}\right)^{2}},
\label{equation13}
\end{equation}
where the mean of $\hat{\mathbf{u}}_{\mathrm{FV}}$ is denoted by $\overline{\mathbf{u}}_{\mathrm{FV}}$.
\subsubsection{Model Validation}\label{S:51}
In this section, we conduct the validation of the proposed model by evaluating its performance on the unseen test dataset and also present the its performance on validation and training datasets. 
\begin{figure}[htbp!]
    \centering
    \subfigure[]{\label{subfig:lab71}\includegraphics[width=0.475\textwidth]{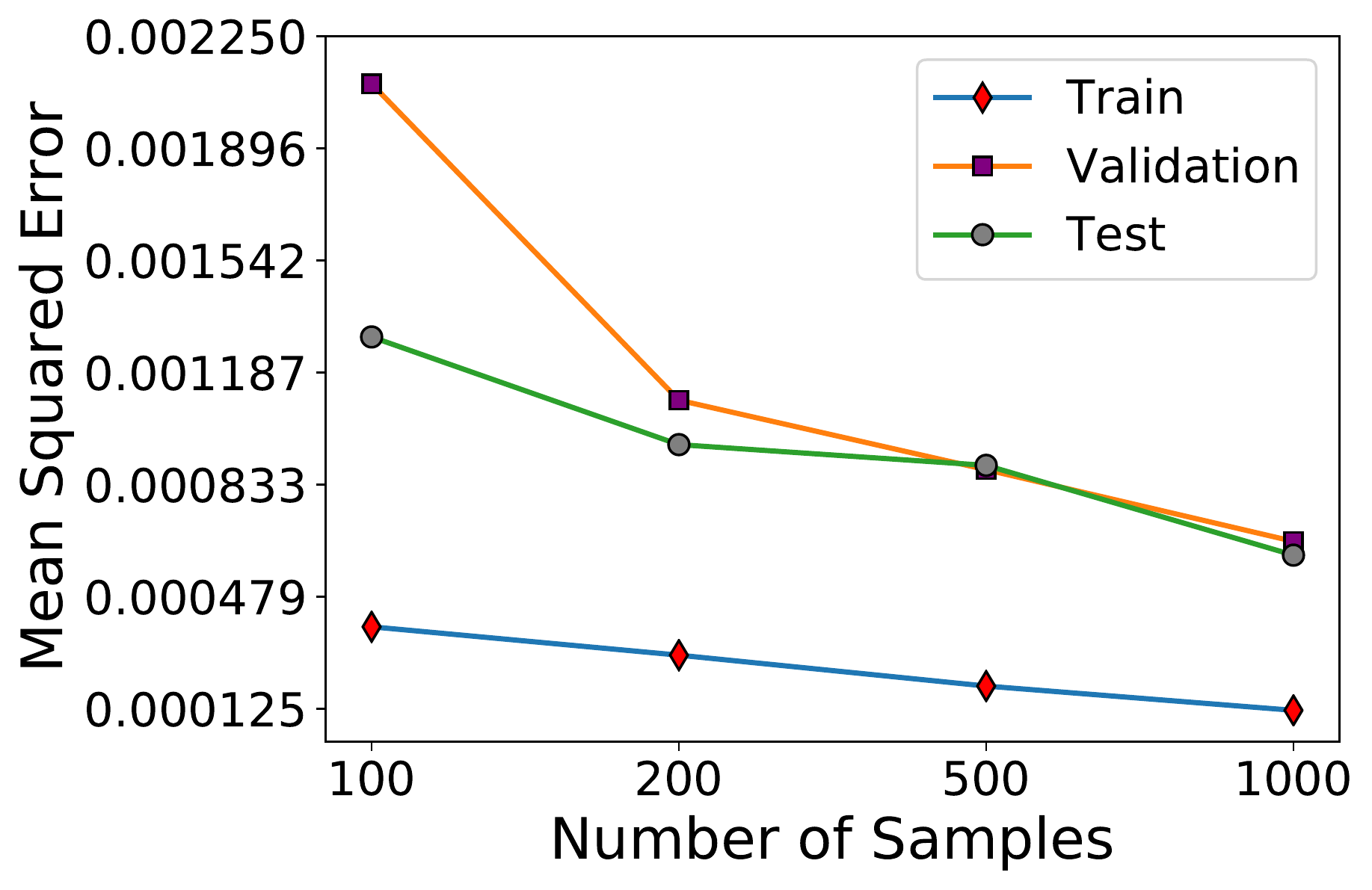}\hspace{4em}} 
    \subfigure[]{\hspace{-3em}\label{subfig:lab72}\includegraphics[width=0.44\textwidth]{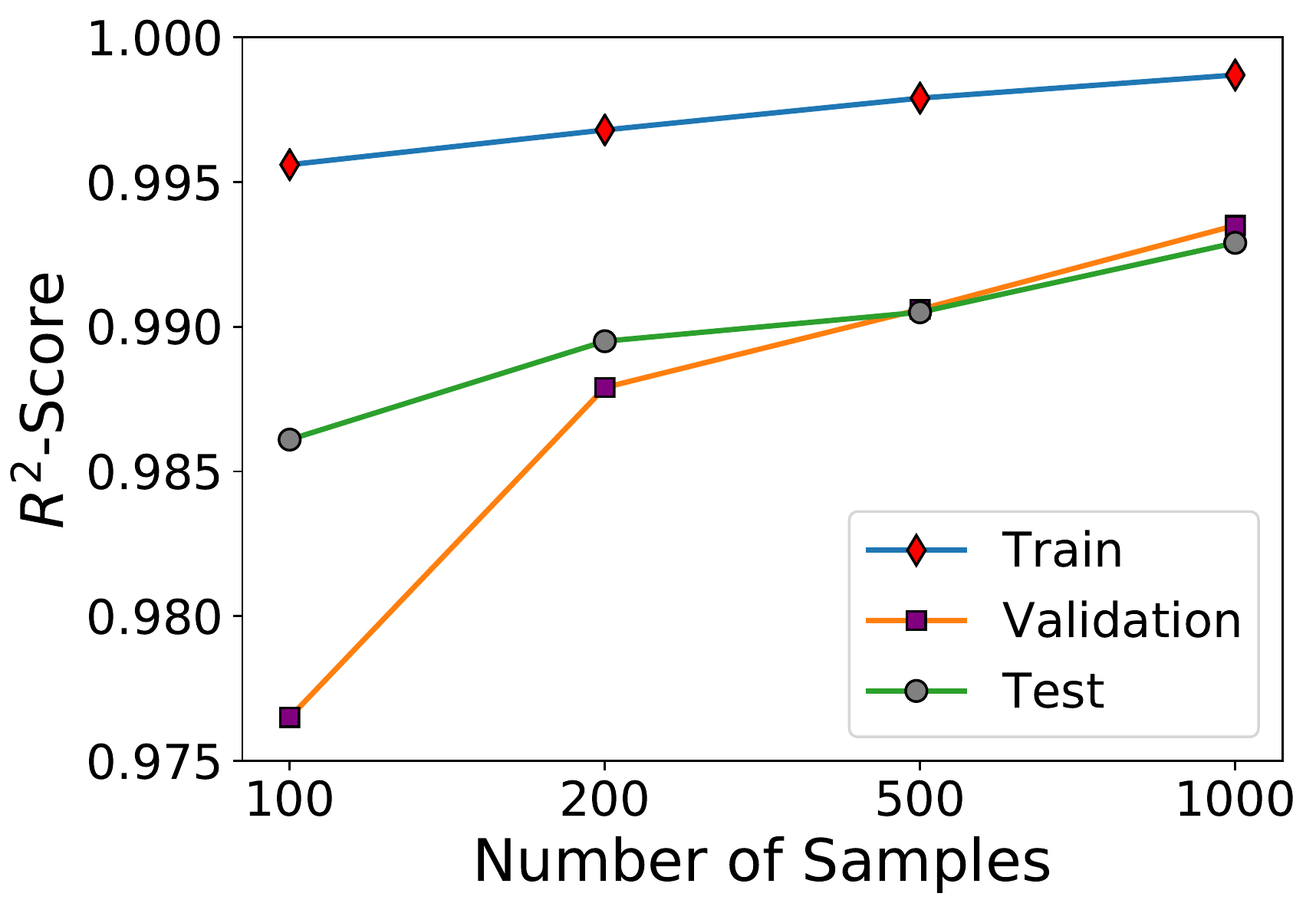}} 
    \caption{\textbf{(a)} Mean-squared error values \textbf{(b)} $R^{2}$-scores on training, validation, and unseen test dataset for different number of samples used for training the surrogate model.}
    \label{fig:7}
\end{figure}
To understand the performance and convergence of the propounded model, the values of the $R^{2}$-scores and mean-squared error on training, validation, and unseen test dataset for different dataset size used for the training the surrogate model are presented in Figure \ref{fig:7}. It is observed from Figure \ref{subfig:lab71} that the values of $R^{2}$- score are consistently above $0.975$ for different training dataset sizes; this establishes the outstanding performance of the proposed framework. Furthermore, for the mean-squared error in \ref{subfig:lab72}, a decrease in error is observed with increase in the the number of training samples, indicating an appropriate training with hardly any overfitting. 
\begin{figure}[htbp!]
\centering
\begin{subfigure}{}
  \centering
  \includegraphics[width=1\linewidth]{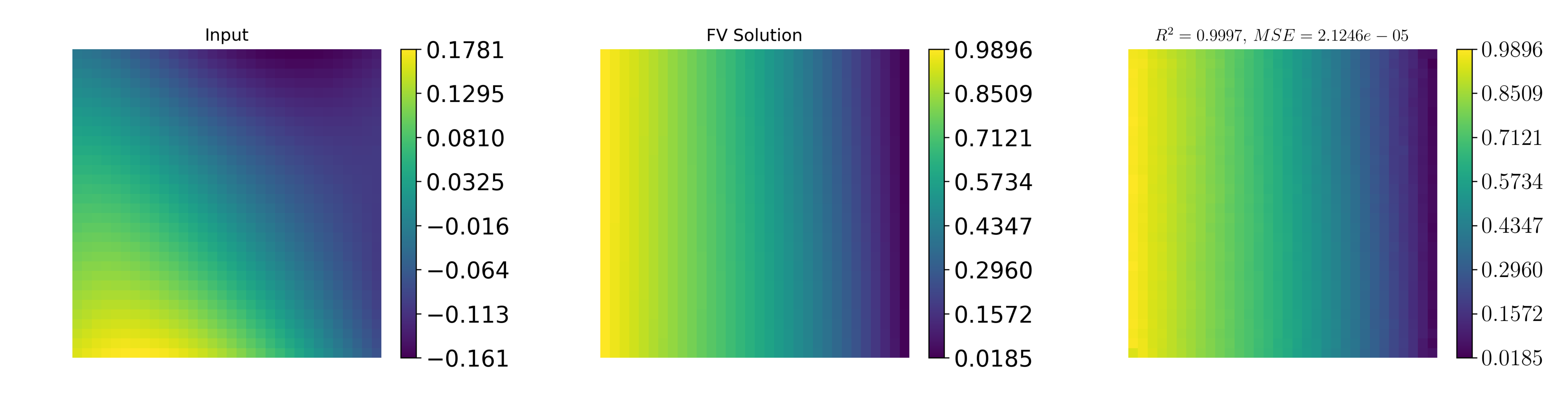}  
\end{subfigure}
\begin{subfigure}{}
  \centering
  \includegraphics[width=1\linewidth]{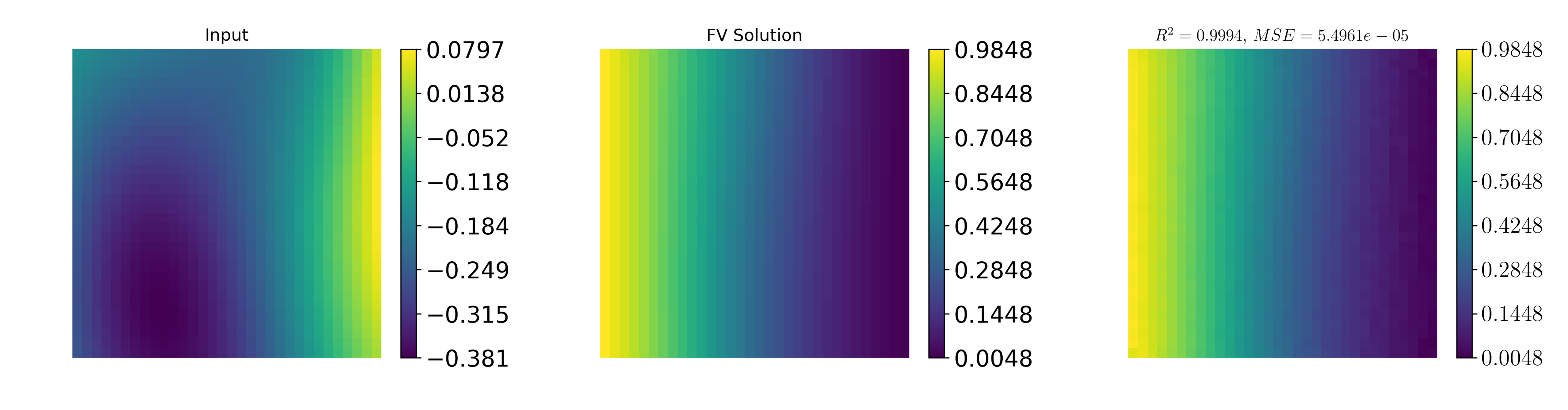}  
\end{subfigure}
\begin{subfigure}{}
  \centering
  \includegraphics[width=1\linewidth]{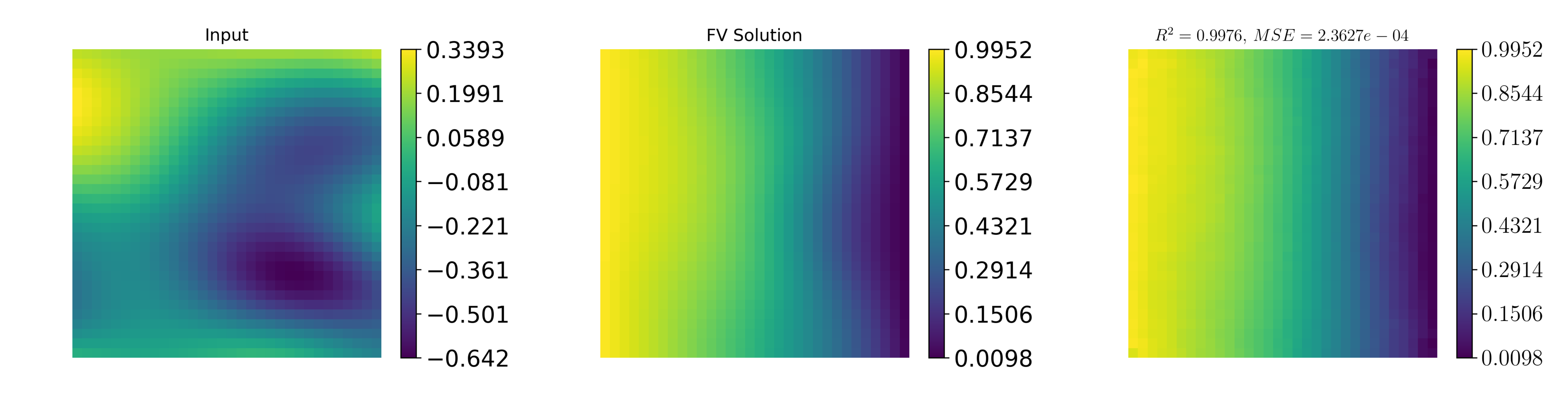}  
\end{subfigure}
\caption{Comparison of predicted PDE solution obtained using the surrogate model with one obtained from FV-solver for a randomly sampled input field from unseen test dataset for surrogate model trained with $1000$ samples. First column shows the input diffusion field, second column shows PDE solution obtained from FVM and third column shows the solution predicted by the surrogate model.}
\label{fig:8}
\end{figure}

We also compare the surrogate model predicted $\hat{\mathbf{u}}$ field on random samples from unseen test dataset with the solution obtained from FVM for surrogate models trained using different training dataset sizes.
\begin{figure}[htbp!]
    \centering
    \subfigure[]{\label{subfig:lab91}\includegraphics[width=0.95\textwidth]{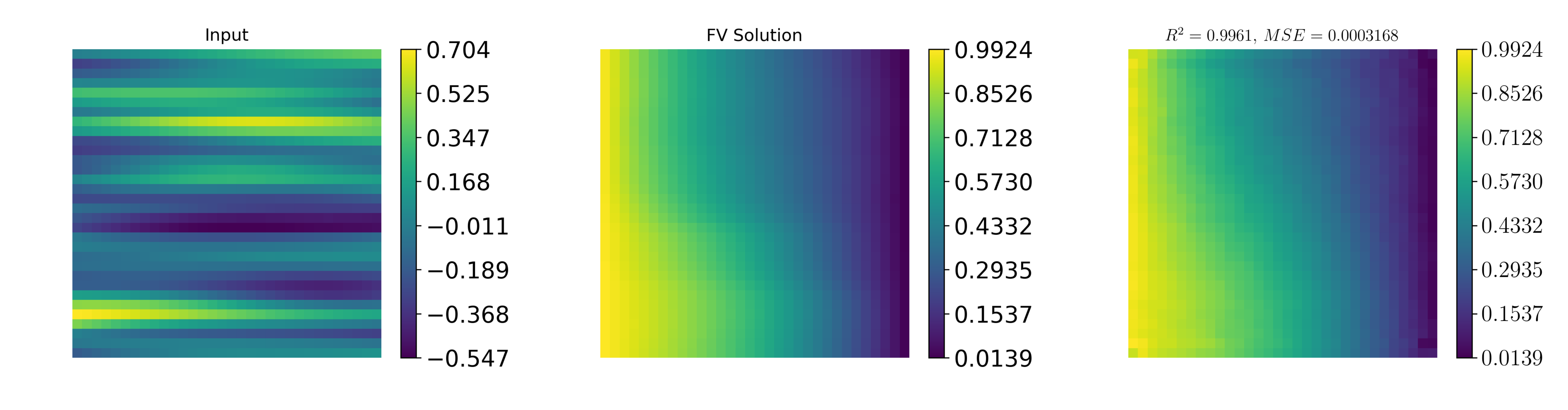}} 
    \subfigure[]{\label{subfig:lab92}\includegraphics[width=0.95\textwidth]{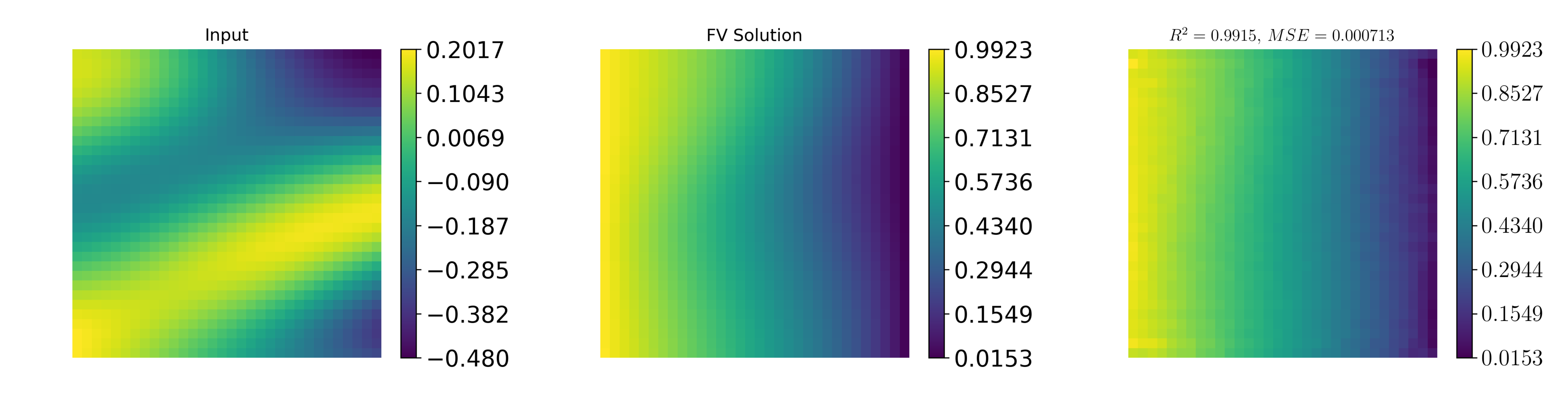}} 
    \subfigure[]{\label{subfig:lab93}\includegraphics[width=0.95\textwidth]{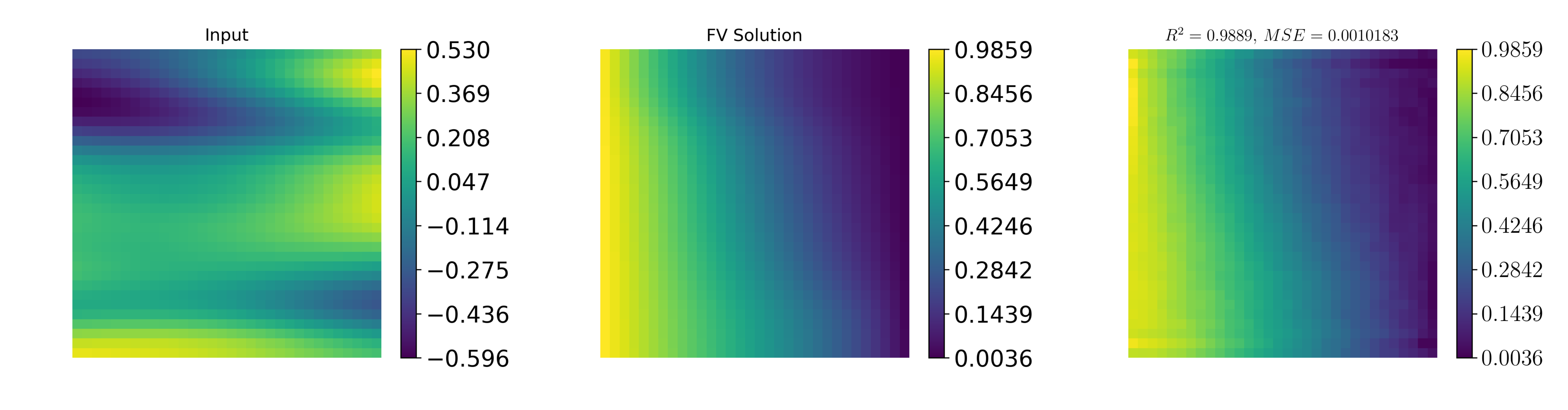}} 
    \caption{Comparison of predicted PDE solution obtained using the surrogate model with one obtained from FV-solver for a randomly sampled input field from unseen test dataset for \textbf{(a)} $500$, \textbf{(b)} $200$, and \textbf{(c)} $100$ training samples. First column shows the input diffusion field, second column shows PDE solution obtained from FVM and third column shows the solution predicted by the surrogate model.}
    \label{fig:9}
\end{figure}
We present the mean squared error and the $R^{2}$-score along with corresponding images for the input random and the predicted and FVM solved output field in Figure \ref{fig:8} and Figure \ref{fig:9}. It could be observed from Figure \ref{fig:8} that with 1000 training samples the surrogate model gives excellent predictive results and error metric values across different lengthscales. Furthermore, from Figure \ref{fig:9}, it is clear that the $\hat{\mathbf{u}}$ field estimates from the surrogate model trained with $500$ training samples are also quite accurate. Moreover, it is noticeable from the results presented for $200$ and $100$ that the predictive accuracy decreases with decreasing training dataset size. However, the predicted $\hat{\mathbf{u}}$ fields still match nicely with the ground truth. Finally, with the presented results, one could confidently state that the surrogate model is able to map the input diffusion coefficient $\hat{\mathbf{a}}$ to solution $\hat{\mathbf{u}}$ accurately. Furthermore, we add that the number of training epochs for successful construction of surrogate model for training dataset sizes $1000$, $500$, $200$, and $100$ are $129$, $133$, $142$, and $145$, respectively. 

\subsubsection{Uncertainty Propagation on arbitrary length scales}\label{S:52}
To evaluate the extrapolative and interpolative predictive capabilities of the proposed framework at arbitrary lengthscales, we assess the ability of developed surrogate model to predict the solution at input random fields which might not even be structurally similar to training dataset samples. Therefore, we subject our framework to the challenge of propagating uncertainty at the following arbitrary lengthscales:
\begin{enumerate}
  \item \textbf{Case 1} - $\ell_{x} = 1.2$ and $\ell_{y} = 1.2$ - Extrapolation.
  \item \textbf{Case 2} - $\ell_{x} = 0.3$ and $\ell_{y} = 0.4$ - Interpolation.
  \item \textbf{Case 3} - $\ell_{x} = \frac{1}{33}$ and $\ell_{y} = \frac{1}{33}$ - Extrapolation
\end{enumerate}
For each of the three cases listed above, we predict the solution $\hat{\mathbf{u}}$ using the surrogate model for $10^{4}$ samples of input random field $\hat{\mathbf{a}}$. We then evaluate and compare the mean and variance of $\hat{\mathbf{u}}$ for the predicted solution and the ground truth obtained from FV-solver. We also assess the performance of the proposed surrogate model by computing the probability density function (pdf) for the solution $\hat{\mathbf{u}}$ at two pairs of selected spatial coordinates and again compare with similar results obtained from ground truth. Finally, we compute and present the $R^{2}$-score and the mean-squared error for each of the three cases. In addition, we conduct all the above-mentioned steps for surrogate models trained with different number of training samples.\\

\begin{figure}[h]
    \centering
    \subfigure[]{\label{subfig:lab101}\includegraphics[width=0.91\textwidth]{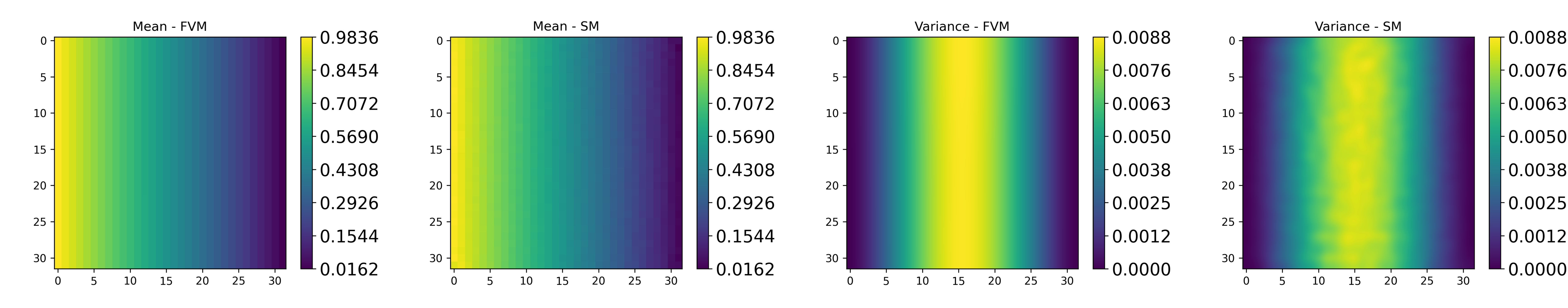}} 
    \subfigure[]{\label{subfig:lab102}\includegraphics[width=0.91\textwidth]{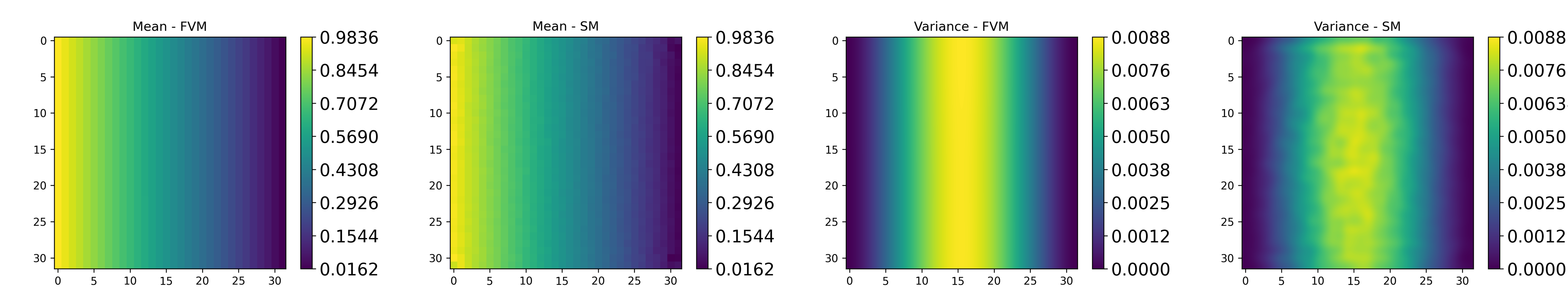}} 
    \subfigure[]{\label{subfig:lab103}\includegraphics[width=0.91\textwidth]{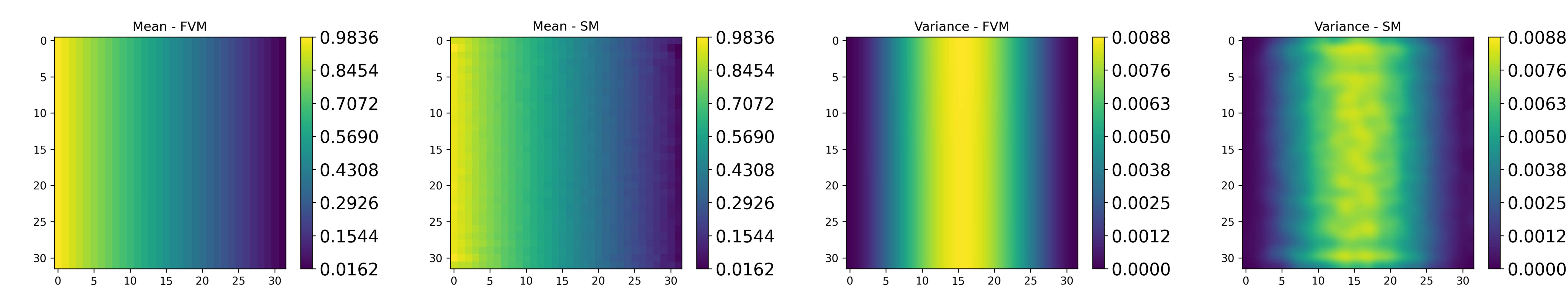}} 
    \caption{ Mean and variance comparison of predicted PDE solution at $(\ell_{x}, \ell_{y}) = (1.2,1.2)$ obtained using the surrogate model with one obtained from FV-solver for \textbf{(a)} $1000$, \textbf{(b)} $500$, and \textbf{(c)} $200$ training samples. First and second columns present the mean of PDE solution obtained from FVM and surrogate model respectively. While the third and fourth columns respectively present the the variance of PDE solution obtained from FVM and surrogate model.}
    \label{fig:10}
\end{figure}

{\textbf{Case 1 Extrapolation at $\ell_{x} = 1.2$ and $\ell_{y} = 1.2$}}:
It could be seen from Figure \ref{fig:10} that, for all the cases with different number of training samples, the mean of the predicted solutions has an excellent match with the ground truth solutions. Also, the values of MSE are very low and the $R^{2}$-score is greater than $0.99$ for all cases. In addition, it is noted that match between the variance of predicted solutions and that of the ground truth is still quite good, and this observation is further strengthened by the low MSE values and higher than $0.975$ values of $R^{2}$-score for different training sizes. It is also observed that the ability of the surrogate model to capture the variance deteriorates with decreasing number of training samples used.

\begin{table}[htbp!]
    \centering
    \begin{tabular}{||c | c | c | c | c ||} 
    \hline
    \multirow{2}{3em}{$\mathcal{D}_{train}$ Size} &\multicolumn{2}{|c|}{Mean} & \multicolumn{2}{|c|}{Variance}\\\cline{2-5}
    & MSE & $R^{2}$-score & MSE  & $R^{2}$-score  \\ [0.5ex] 
    \hline
    $1000$ & 1.9216e-05 & 0.9998 & 4.0161e-08 & 0.9958  \\ 
    \hline
    $500$ & 5.6788e-05 & 0.9993 & 1.1456e-07 & 0.9879  \\ 
    \hline
    $200$ & 7.1740e-05 & 0.9991 & 2.5668e-07 & 0.9729  \\ 
    \hline
    \end{tabular}
    \caption{MSE and $R^{2}$-score for mean and variance of PDE solution for $\ell_{x} = 1.2$ and $\ell_{y} = 1.2$.}
    \label{tab:1}
\end{table}

\begin{figure}[H]
    \centering
    \subfigure[]{\label{subfig:lab111}\includegraphics[width=0.45\textwidth]{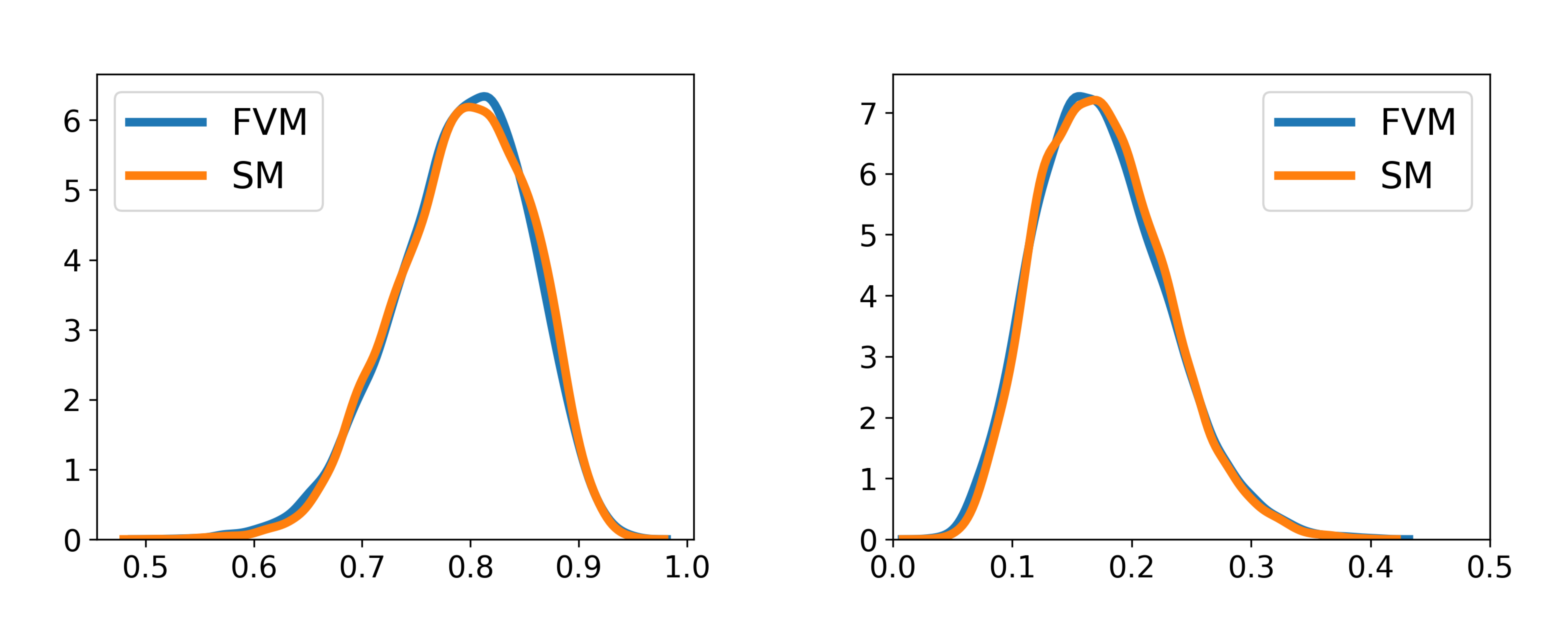}} 
    \subfigure[]{\label{subfig:lab112}\includegraphics[width=0.45\textwidth]{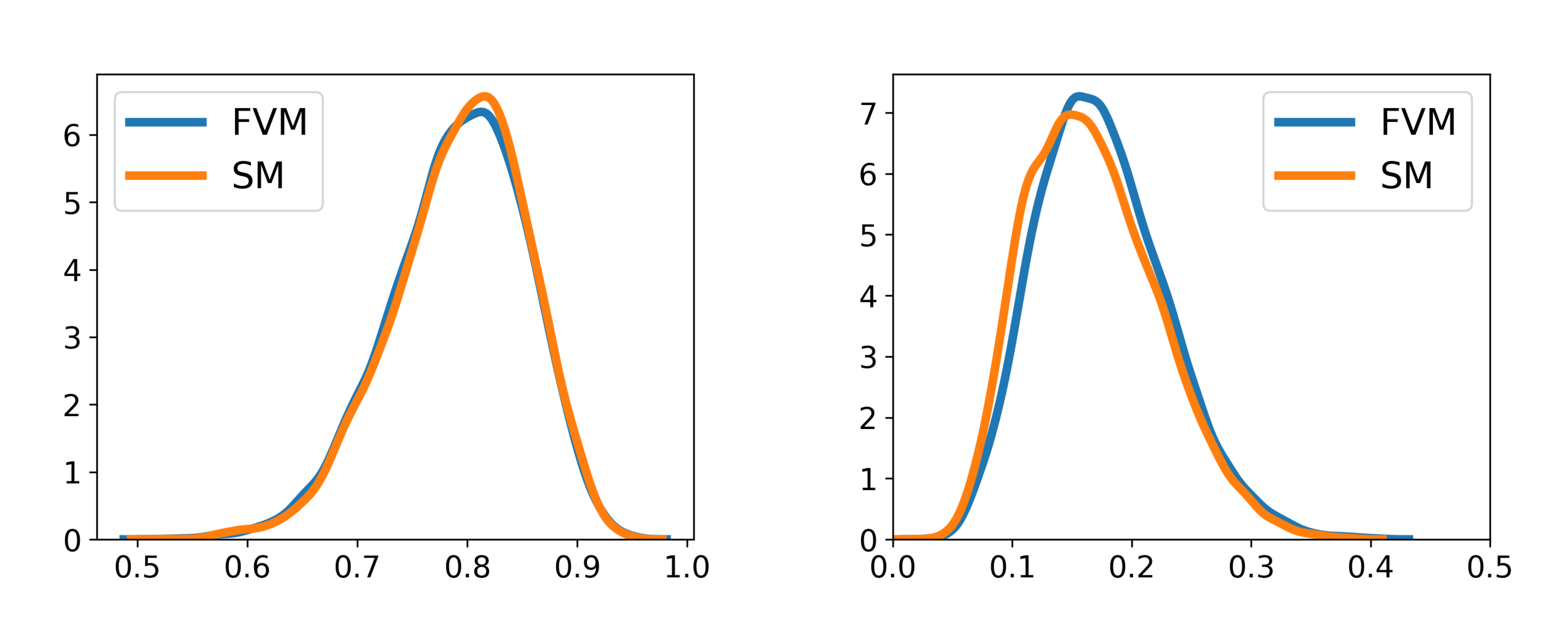}} 
    \subfigure[]{\label{subfig:lab113}\includegraphics[width=0.45\textwidth]{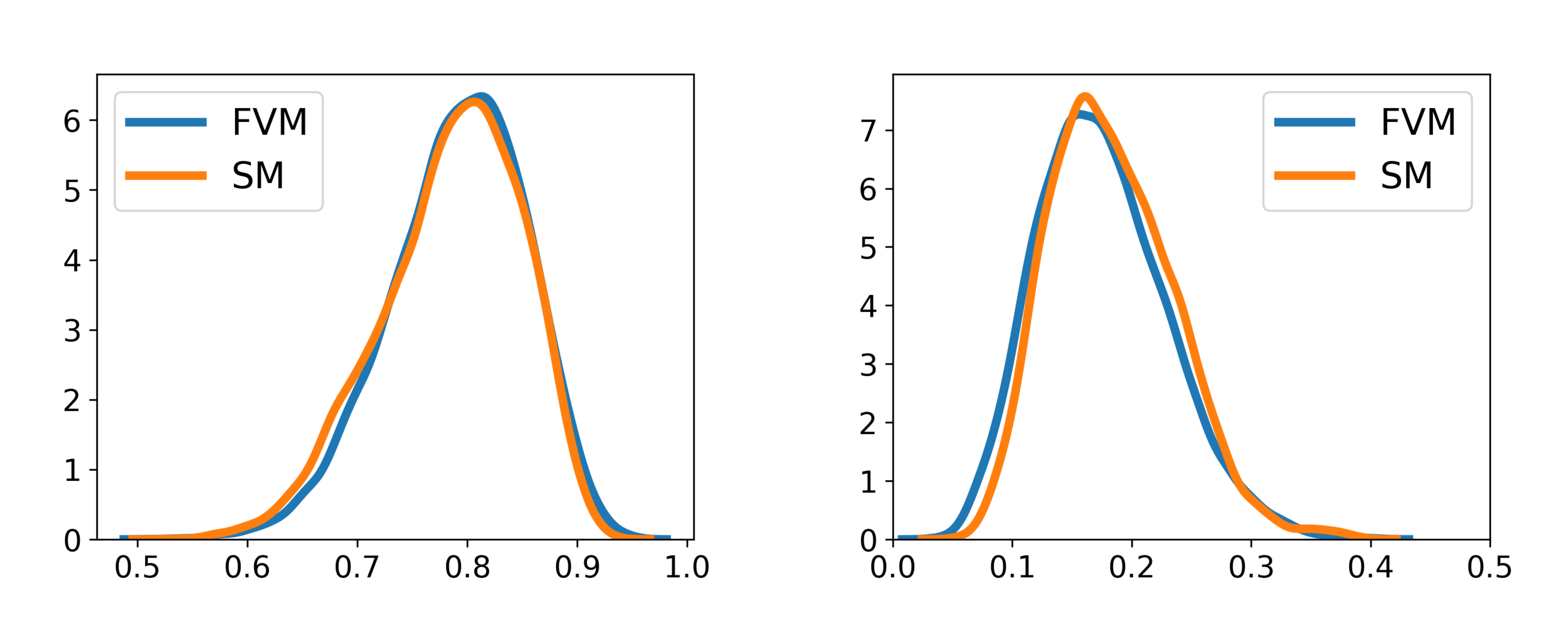}} 
    \caption{ PDE solution density at $\mathbf{x}_{1} = (0.203125, 0.203125)$ and $\mathbf{x}_{2} = (0.828125, 0.703125)$ at $(\ell_{x}, \ell_{y}) = (1.2,1.2)$ for \textbf{(a)} $1000$, \textbf{(b)} $500$, and \textbf{(c)} $200$ training samples. Left plot in each case is for $\mathbf{x}_{1} = (0.203125, 0.203125)$ and right plot is for $\mathbf{x}_{2} = (0.828125, 0.703125)$.}
    \label{fig:11}
\end{figure}
Finally, we compare the pdf of the solution obtained from the surrogate model and the FV-solver at two spatial locations, namely, $\mathbf{x}^{(1)} = (0.203125, 0.203125)$ and $\mathbf{x}^{(2)} = (0.828125, 0.703125)$ for surrogate models trained with different number of training samples. The results are presented in Figure \ref{fig:11}. It is evident from the figure that the computed pdf at both $\mathbf{x}^{(1)}$ and $\mathbf{x}^{(2)}$ for the surrogate model matches quite well with the ground truth obtained using FVM for all the training dataset sizes. This is attributed to that the fact that the solution at larger lengthscales is very smooth and thus, it is easier for the surrogate model to learn and predict.\\

{\textbf{Case 2 Interpolation at $\ell_{x} = 0.3$ and $\ell_{y} = 0.4$}}:
We now assess the interpolative capabilities of our proposed framework by propagating the uncertainty for the lengthscales $\ell_{x} = 0.3$ and $\ell_{y} = 0.4$. 
\begin{figure}[h]
    \centering
    \subfigure[]{\label{subfig:lab121}\includegraphics[width=0.94\textwidth]{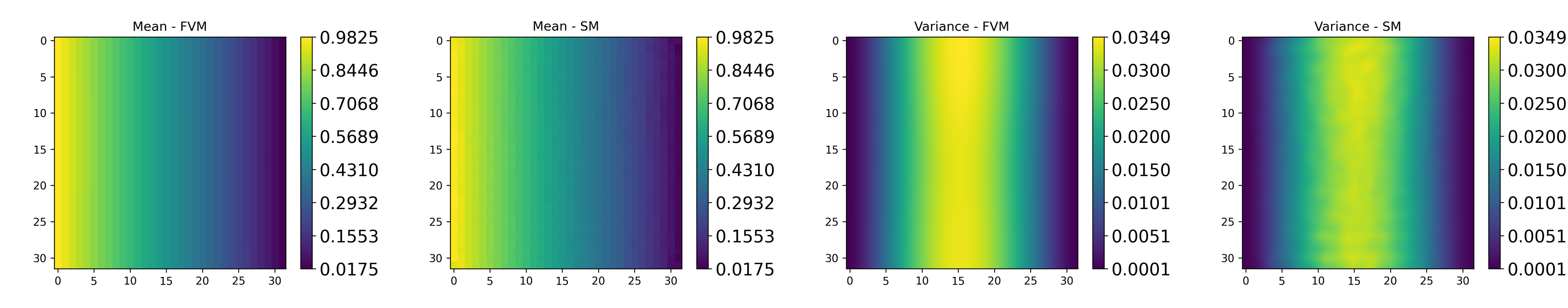}} 
    \subfigure[]{\label{subfig:lab122}\includegraphics[width=0.94\textwidth]{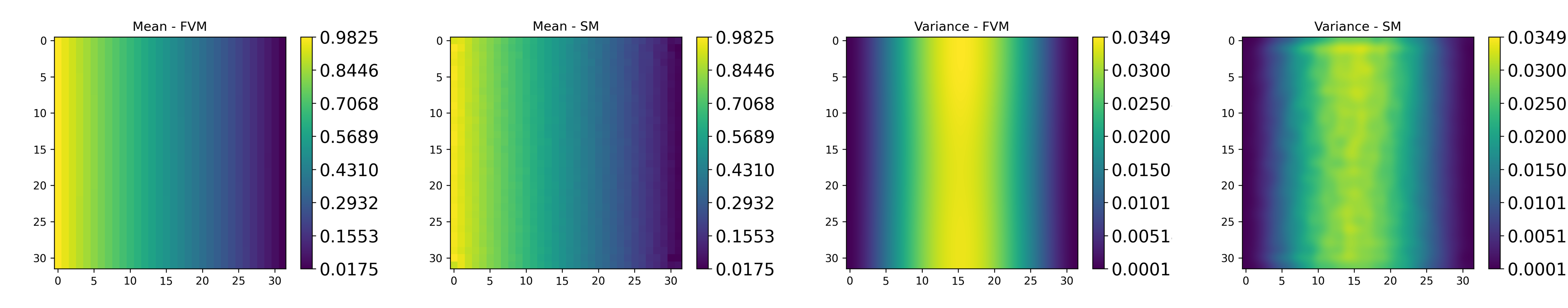}} 
    \subfigure[]{\label{subfig:lab123}\includegraphics[width=0.94\textwidth]{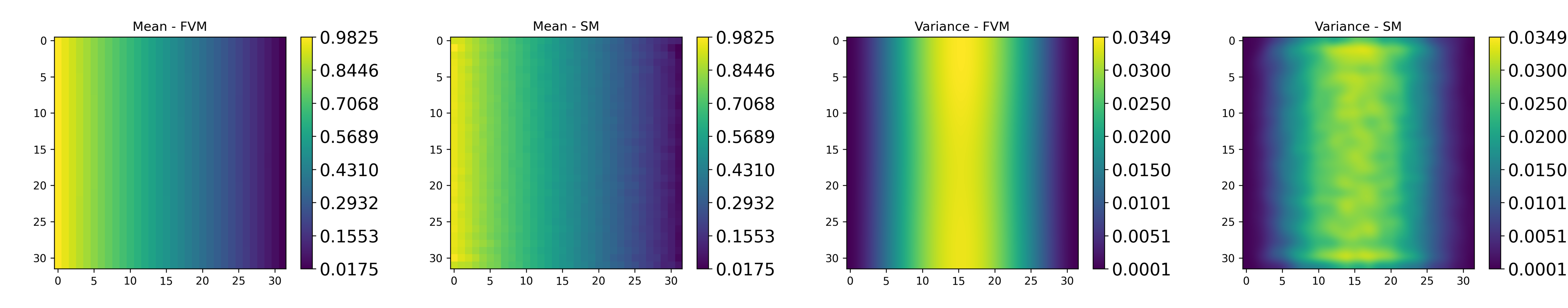}} 
    \caption{ Mean and variance comparison of predicted PDE solution at $(\ell_{x}, \ell_{y}) = (0.3,0.4)$ obtained using the surrogate model with one obtained from FV-solver for \textbf{(a)} $1000$, \textbf{(b)} $500$, and \textbf{(c)} $200$ training samples. First and second columns present the mean of PDE solution obtained from FVM and surrogate model respectively. While the third and fourth columns respectively present the the variance of PDE solution obtained from FVM and surrogate model.}
    \label{fig:12}
\end{figure}
Similar to \textbf{Case I}, it is observed that the surrogate model is able to accurately capture the mean and variance of $\hat{\mathbf{u}}$ (see Figure \ref{fig:12}). Also, the MSE and $R^{2}$- score between the statistics, i.e., mean and variance, estimated through the surrogate model and the statistics obtained from the FV-solution presented in Table \ref{tab:2} further reinforce the observations from Figure \ref{fig:12}.
\begin{table}[htbp!]
    \centering
    \begin{tabular}{|| c | c | c | c | c ||} 
    \hline
    \multirow{2}{3em}{$\mathcal{D}_{train}$ Size} &\multicolumn{2}{|c|}{Mean} & \multicolumn{2}{|c|}{Variance}\\\cline{2-5}
    & MSE & $R^{2}$-score & MSE  & $R^{2}$-score  \\ [0.5ex] 
    \hline
    $1000$ & 2.2243e-05 & 0.9997 & 4.4934e-06 & 0.9677  \\ 
    \hline
    $500$ & 9.7306e-05 & 0.9987 & 3.8803e-06 & 0.9721  \\ 
    \hline
    $200$ & 6.40434e-05& 0.9991 & 1.4649e-05 & 0.8949  \\ 
    \hline
    \end{tabular}
    \caption{MSE and $R^{2}$-score for mean and variance of PDE solution for the lengthscale pair $\ell_{x} = 0.3$ and $\ell_{y} = 0.4$.}
    \label{tab:2}
\end{table}

\begin{figure}[H]
    \centering
    \subfigure[]{\label{subfig:lab131}\includegraphics[width=0.45\textwidth]{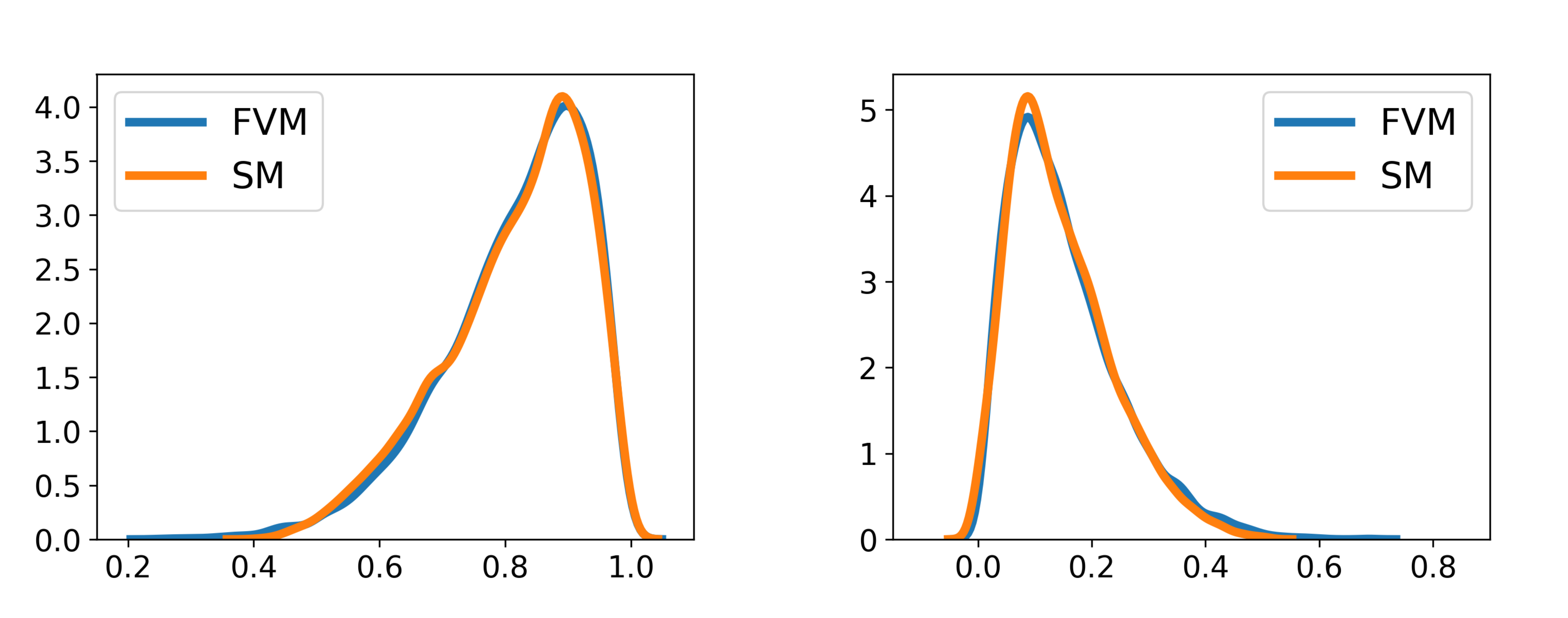}} 
    \subfigure[]{\label{subfig:lab132}\includegraphics[width=0.45\textwidth]{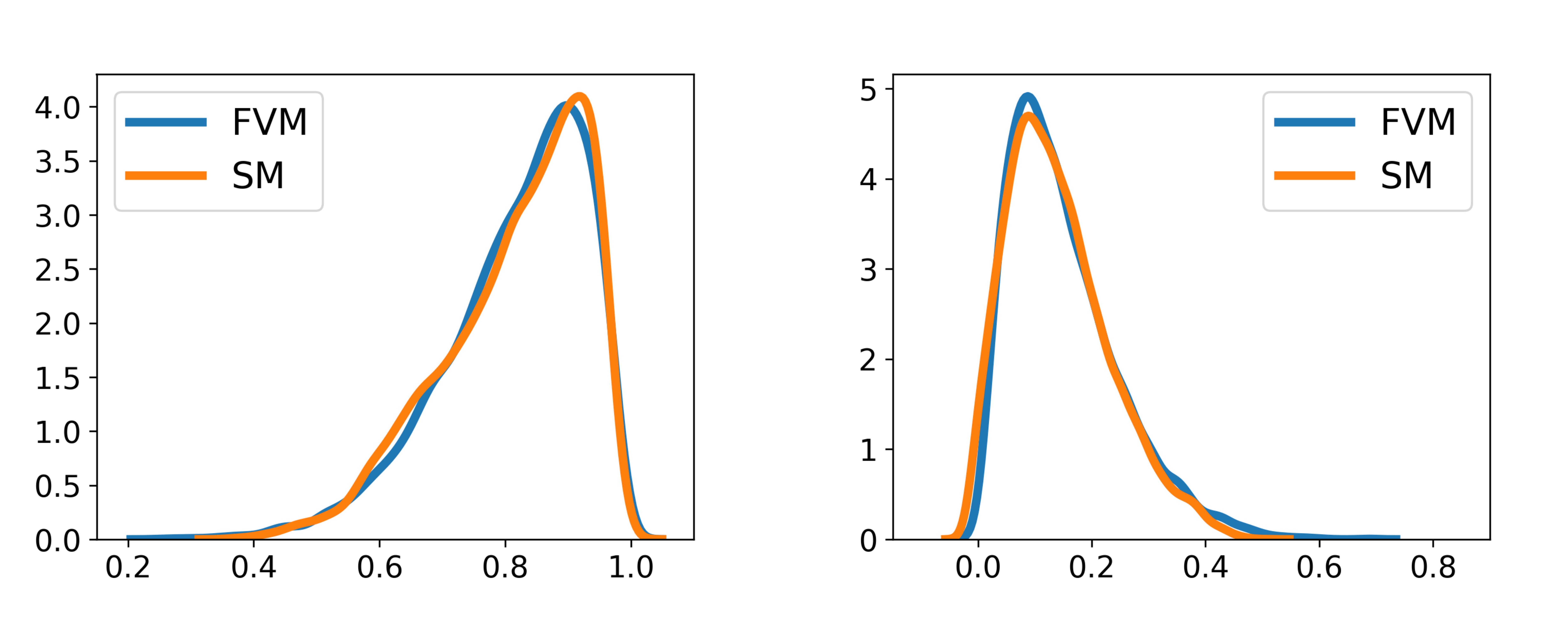}} 
    \subfigure[]{\label{subfig:lab133}\includegraphics[width=0.45\textwidth]{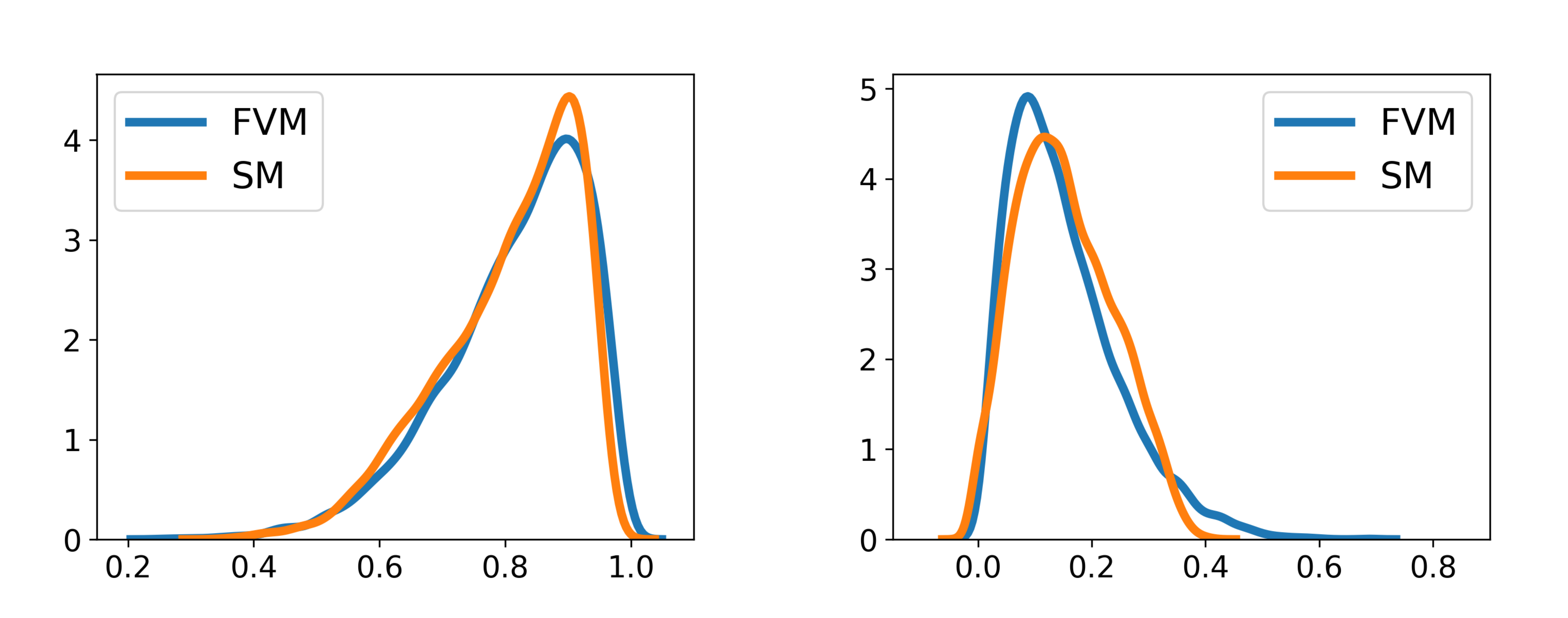}} 
    \caption{ PDE solution density at $\mathbf{x}_{1} = (0.171875, 0.203125)$ and $\mathbf{x}_{2} = (0.859375, 0.921875)$ at $(\ell_{x}, \ell_{y}) = (0.3,0.4)$ for \textbf{(a)} $1000$, \textbf{(b)} $500$, and \textbf{(c)} $200$ training samples. Left plot in each case is for $\mathbf{x}_{1} = (0.171875, 0.203125)$ and right plot is for $\mathbf{x}_{2} = (0.859375, 0.921875)$.}
    \label{fig:13}
\end{figure}
We also compare the pdf of the solution obtained from the surrogate model and the FV- solver at two spatial grid points, in this case, $\mathbf{x}_{1} = (0.171875, 0.203125)$ and $\mathbf{x}_{2} = (0.859375,0.921875)$ for model trained with different sized training dataset. The results are presented in Figure \ref{fig:13}, and the results indicate an excellent match between pdf computed using the surrogate model to that from the FV-solver. Also, the disparity between the surrogate computed and the ground truth evaluated pdf increases with decrease in the training dataset size.\\

\begin{figure}[H]
    \centering
    \subfigure[]{\label{subfig:lab141}\includegraphics[width=0.9\textwidth]{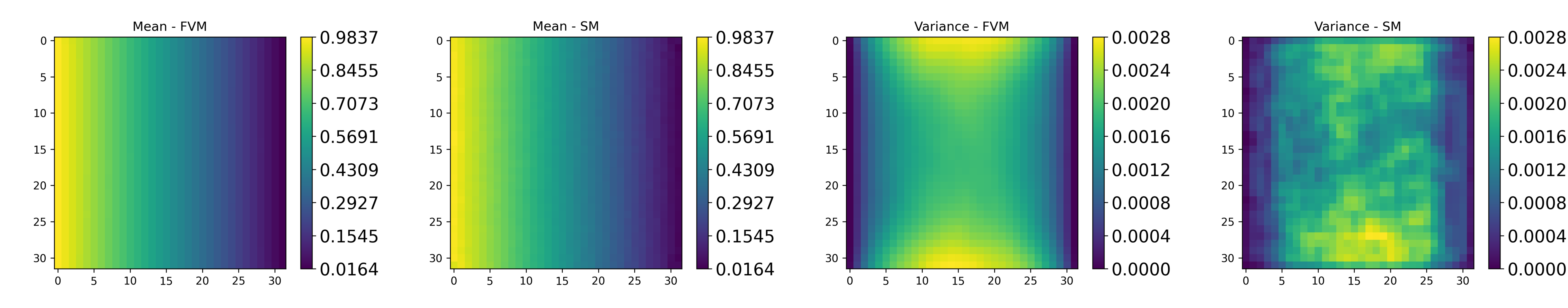}} 
    \subfigure[]{\label{subfig:lab142}\includegraphics[width=0.9\textwidth]{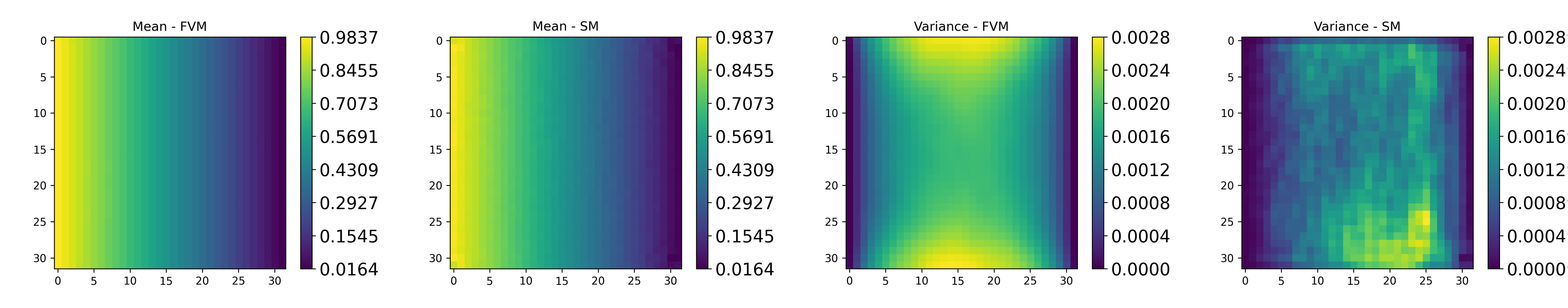}} 
    \subfigure[]{\label{subfig:lab143}\includegraphics[width=0.9\textwidth]{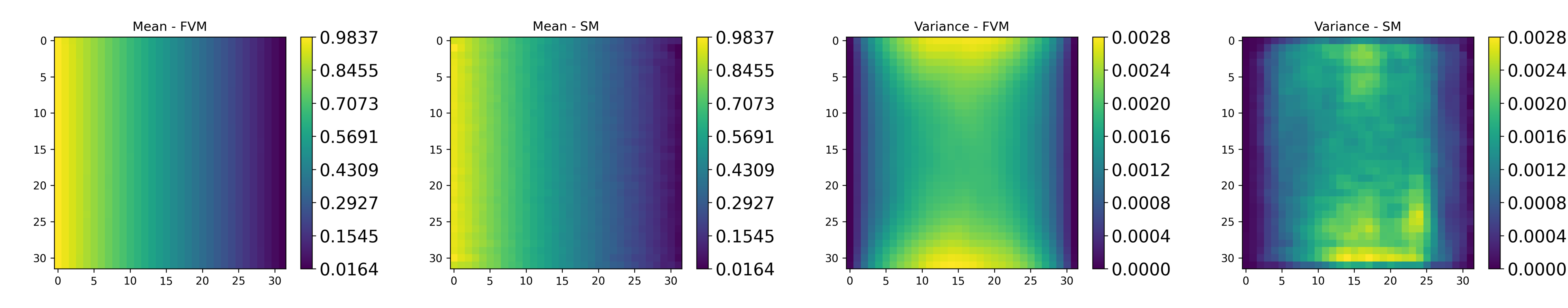}} 
    \caption{ Mean and variance comparison of predicted PDE solution at $(\ell_{x}, \ell_{y}) = \left(\frac{1}{33}, \frac{1}{33}\right)$ obtained using the surrogate model with one obtained from FV-solver for \textbf{(a)} $1000$, \textbf{(b)} $500$, and \textbf{(c)} $200$ training samples. First and second columns present the mean of PDE solution obtained from FVM and surrogate model respectively. While the third and fourth columns respectively present the the variance of PDE solution obtained from FVM and surrogate model.}
    \label{fig:14}
\end{figure}

{\textbf{Case 3 Extrapolation at $\ell_{x} = \frac{1}{33}$ and $\ell_{y} = \frac{1}{33}$}:}
Finally, we conduct uncertainty propagation for the extrapolation case of lengthscale pair $(\ell_{x}, \ell_{y}) = \left(\frac{1}{33}, \frac{1}{33}\right)$, which is smaller than the size of the FV-cell. It is clear from Figure \ref{fig:14} that the surrogate model struggles to estimate variance of the PDE solution at all sizes of dataset used for training the surrogate model. But this is expected because, as Tripathy and Bilionis \cite{tripathy2018deep} pointed out, the solution becomes inherently unsmooth at smaller lengthscales and therefore, it becomes very difficult for the proposed framework to predict the solution correctly. It is also observed that the estimation of the mean of  $\hat{\mathbf{u}}$ field by surrogate model with all the training sizes has a low MSE value and a $R^{2}$-score greater than $0.99$.

\begin{figure}[h]
    \centering
    \subfigure[]{\label{subfig:lab151}\includegraphics[width=0.45\textwidth]{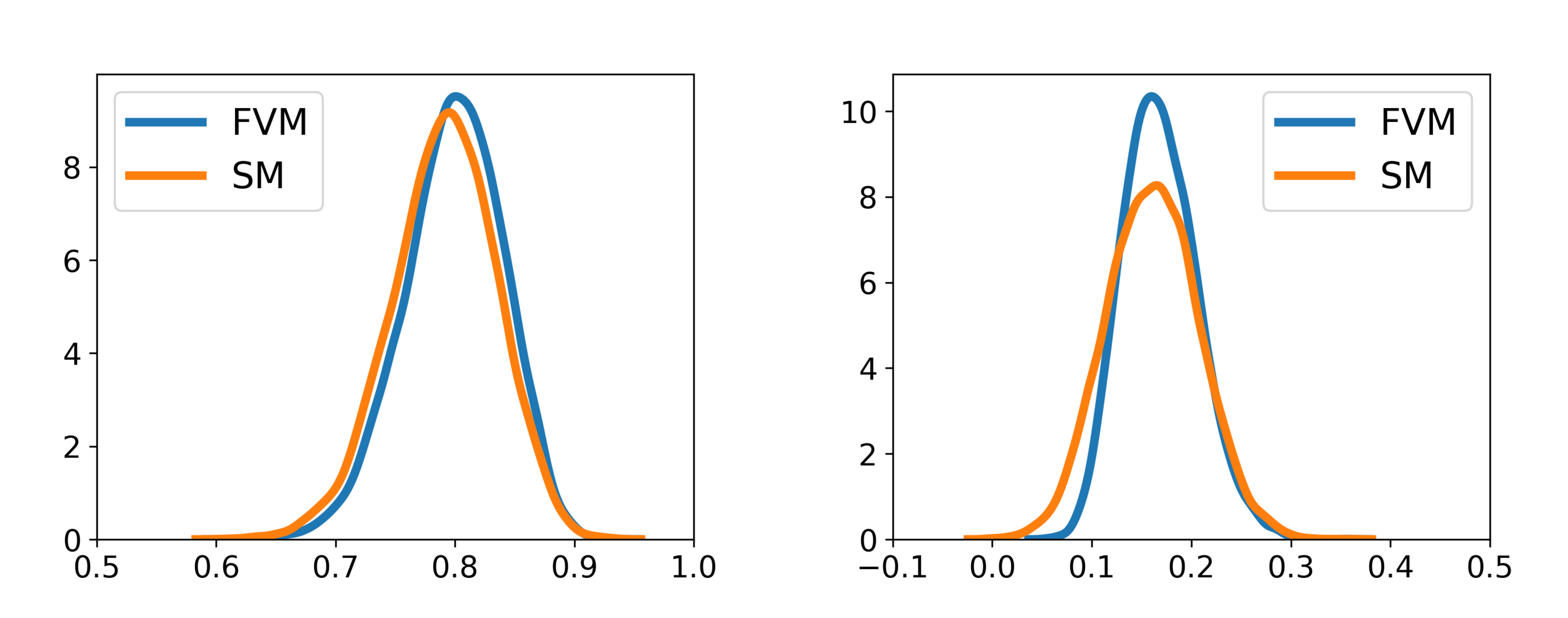}} 
    \subfigure[]{\label{subfig:lab152}\includegraphics[width=0.45\textwidth]{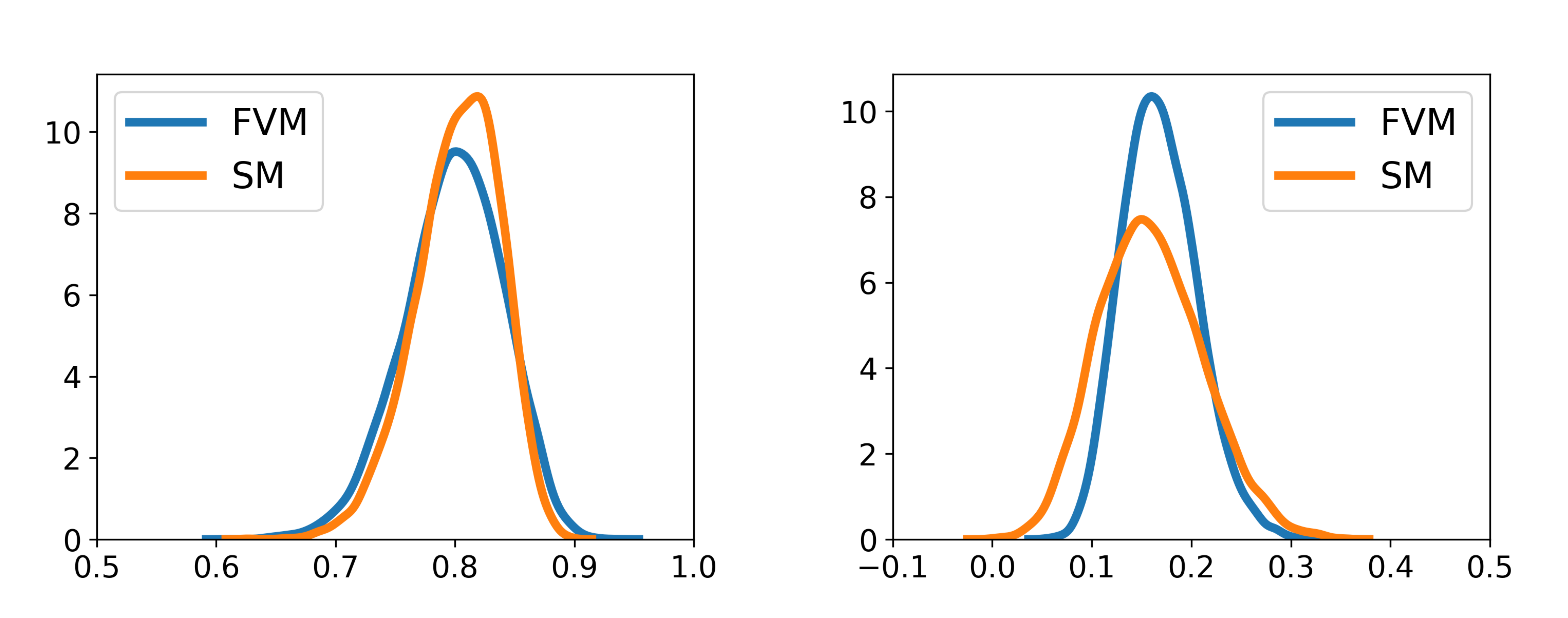}} 
    \subfigure[]{\label{subfig:lab153}\includegraphics[width=0.45\textwidth]{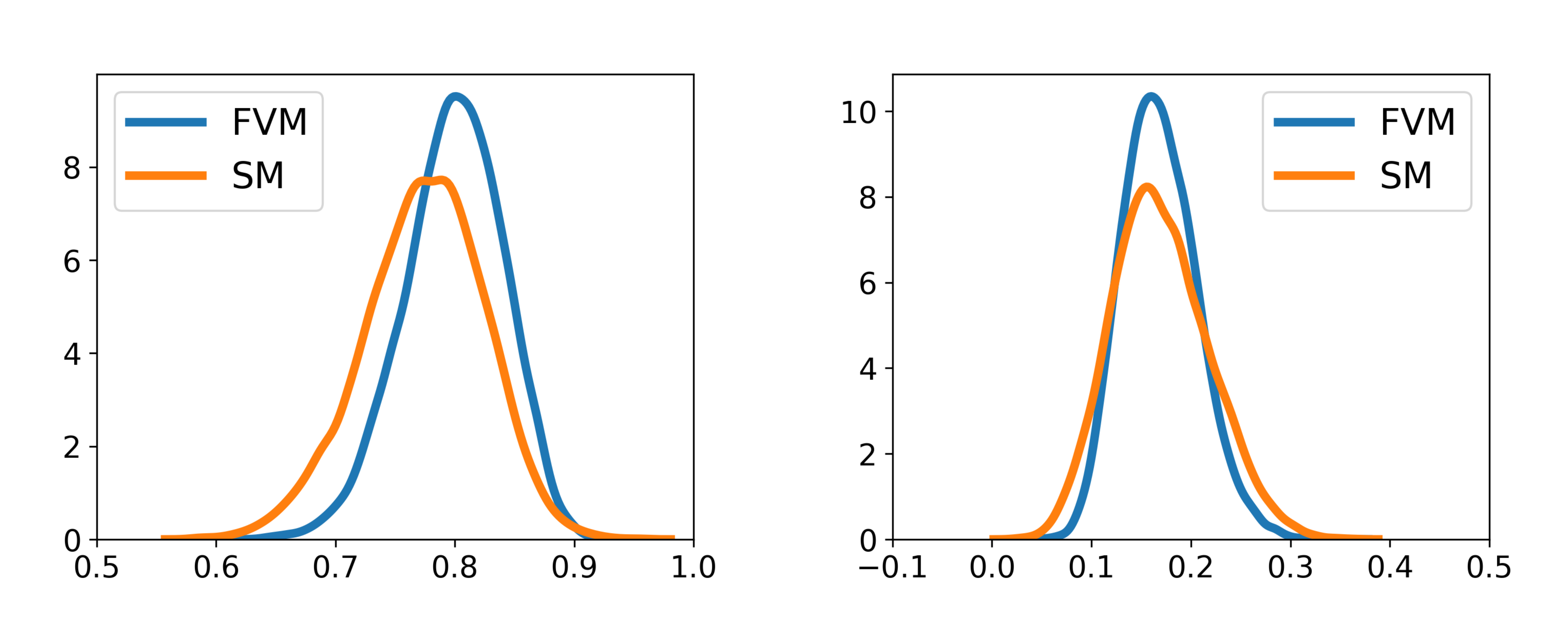}} 
    \caption{ PDE solution density at $\mathbf{x}_{1} = (0.203125, 0.203125)$ and $\mathbf{x}_{2} = (0.828125, 0.703125)$ at $(\ell_{x}, \ell_{y}) = \left(\frac{1}{33}, \frac{1}{33}\right)$ for \textbf{(a)} $1000$, \textbf{(b)} $500$, and \textbf{(c)} $200$ training samples. Left plot in each case is for $\mathbf{x}_{1} = (0.203125, 0.203125)$ and right plot is for $\mathbf{x}_{2} = (0.828125, 0.703125)$.}
    \label{fig:15}
\end{figure}
 The comparative inability of the surrogate model to predict solution at smaller than FV-cell size extrapolative lengthscale is further made evident by the relatively larger mismatch between pdf computed from the surrogate solution and that computed from the solution obtained from FV-solver (see  Figure \ref{fig:15}). Nonetheless, it is worthwhile to note that even for this case, the predicted pdfs still provide a reasonable trend of the response uncertainty.

\section{Conclusions}\label{S:6}
In this study, we propose a novel deep learning architecture for surrogate modeling and uncertainty quantification of high dimensional PDE systems. In the proposed approach, we exploits the superior expressive capability of capsules and adapt it into an image-to-image regression encoder-decoder network.
Overall, the proposed architecture is highly efficient, robust, and works with sparse dataset. \par 
The performance of the proposed approach is evaluated on an elliptic SPDE with a dimensionality of $1024$ and removed assumptions on lengthscale of input random field. The proposed framework does an excellent job in predicting the solution at unseen input random fields at arbitrary lengthscales regardless of it being an extrapolation or interpolation task; this claim is based upon the low values and high $R^{2}$- score for different tasks that the model is subjected to. Also, the approach performs well for quantifying the uncertainty propagation from the input field to output field. Moreover, the developed model requires lesser number of training samples and training epochs (not more than $145$ for all the different training dataset sizes) for construction of a successful surrogate model. Also, the performance of the proposed is found reasonable as the number of training samples is decreased and the model is found robust in the convergence studies. Furthermore, it is also found that the developed CapsNet based framework requires minimal hyperparameter tuning for appropriate learning. \par
However, it is also noted that the developed framework's performance deteriorates when the lengthscales becomes very small, but at the same time it is understandable as the solutions become quite unsmooth and irregular. Also, at present the proposed architecture is frequentist in nature; future work in this direction will involve developing a Bayesian surrogate with the same architecture.

\section*{Acknowledgement}
SC and AT acknowledge the open-source code provided by Tripathy and Bilionis \cite{tripathy2018deep} and Xifeng Guo.

\end{document}